\documentclass[journal]{IEEEtran}

\usepackage{amsmath,amssymb,amsfonts}
\usepackage{algorithmic}
\usepackage{graphicx}
\usepackage{textcomp}
\usepackage{xcolor}
\usepackage{soul}
\usepackage{cancel}
\usepackage{cite}

\ifCLASSINFOpdf
\else
\fi
\begin{document}

\title{Probabilistic approach to feedback control enhances multi-legged locomotion on rugged landscapes 
}

\author{Juntao~He,~\IEEEmembership{Student member,~IEEE,}
        Baxi~Chong,~\IEEEmembership{Student member,~IEEE,}
        Jianfeng~Lin,~\IEEEmembership{Student member,~IEEE,}
        Zhaochen~Xu,
        Hosain Bagheri,
        Esteban~Flores,
        Daniel I.~Goldman
\thanks{J. He is with Institute for Robotics and Intelligent Machines, Atlanta,
GA, 30332 USA e-mail: \textit{jhe391@gatech.edu}}
\thanks{B. Chong, J. Lin, Z. Xu, H. Bagheri, D. I. Goldman are with School of Physic, Georgia Institute of Technology, Atlanta,
GA, 30332 USA e-mail: \textit{\{{bchong9, jianf.lin, zxu699, hbagheri}\}@gatech.edu} and \textit{daniel.goldman@physics.gatech.edu} }
\thanks{E. Flores is with School of Mechanical Engineering, Georgia Institute of Technology, Atlanta,
GA, 30332 USA e-mail: \textit{eflores36@gatech.edu}}
\thanks{Corresponding author: Juntao He.}
}

\maketitle

\begin{abstract}
Achieving robust legged locomotion on complex terrains poses challenges due to the high uncertainty in robot-environment interactions. Recent advances in bipedal and quadrupedal robots demonstrate good mobility on rugged terrains but rely heavily on sensors for stability due to low static stability from a high center of mass and a
narrow base of support\cite{ijspeert2023integration}. We hypothesize that a multi-legged robotic system can leverage morphological redundancy from additional legs to minimize sensing requirements when traversing challenging terrains. Studies suggest \cite{chong2023science,chong2023pnas} that a multi-legged system with sufficient legs can reliably navigate noisy landscapes without sensing and control, albeit at a low speed of up to 0.1 body lengths per cycle (BLC). However, the feedback control framework to enhance speed of multi-legged robots on challenging terrains remains underexplored due to diverse environmental interactions. Such complexity makes it difficult to identify the key parameters to control in these high-degree-of-freedom systems. Here, in laboratory and field experiments we demonstrate that a vertical body undulation wave helps to mitigate environmental disturbances affecting robot speed; these observations are supported by a probabilistic model. Using such insights, we introduce a control framework which monitors foot-ground contact patterns on rugose landscapes using binary foot-ground contact sensors to estimate terrain rugosity. The controller adjusts the vertical body wave based on the deviation of the limb's averaged actual-to-ideal foot-ground contact ratio, achieving a significant enhancement of up to 0.235 BLC on rugose laboratory terrain. We observed a 50\% to 60\% increase in speed and a 30\% to 50\% reduction in speed variance compared to the open-loop controller. Additionally, the controller operates in complex terrains outside the lab, including pine straw, robot-sized rocks, mud, and leaves.
\end{abstract}


%
\IEEEpeerreviewmaketitle

\section{Introduction}
\IEEEPARstart{L}{egged} robots offer a compelling alternative to wheeled robots, particularly when navigating unstructured terrain. Recent advances in few-legged robots demonstrate their adaptability to diverse and unstructured terrains. Bipedal robots excel at obstacle avoidance and stair climbing \cite{shamsah2023integrated,yagi1999biped,siekmann2021blind}, but their static instability necessitates substantial effort to maintain balance in an upright posture. Disruptions in body or leg trajectories can lead to instability \cite{hohn2009probabilistic,wight2008introduction}, limiting mobility on highly uneven terrains. Quadruped robots, known for greater stability, perform well on challenging terrains like snow, wet moss, mud, and rocky surfaces \cite{choi2023learning,lee2020learning,miki2022learning}. However, many advanced robotic systems rely on force sensors, accelerometers \cite{bloesch2013state,boaventura2012dynamic,lee2020development,hutter2016anymal}, cameras, and LiDAR \cite{agarwal2023corl,agrawal2022vision,wisth2022vilens,bellicoso2018advances,dang2020graph} for analyzing foot-environment interactions and estimating terrain geometry. However, not all legged robots require these systems. Systems with high static stability and morphological redundancy, such as hexapods \cite{galloway2010x,saranli2001rhex} or robots with more legs \cite{chong2023science,chong2023pnas,chong2022general}, often traverse rugged terrain effectively with simpler sensing and control frameworks. These simpler systems minimize sensory complexity while still achieving reliable locomotion. 

Recent studies \cite{chong2023science,chong2023pnas} indicate that a serially connected multi-legged robotic system with high static stability and morphological redundancy can reliably traverse noisy landscapes without requiring sensing and control. These systems successfully transport between two points on rough terrain without feedback control. However, their speed on such terrain with an open-loop controller is relatively low and depends on multiple legs to maintain it. 

To address these limitations and improve the effectiveness of multi-legged locomotion in complex environments, designing a feedback control framework is essential. Specifically, designing an simple feedback controller to enhance the robot's performance with minimal sensing is a challenging yet fascinating problem. To the best of our knowledge, feedback control for such multi-legged robotic systems remains a challenging area of research, particularly due to their high degrees of freedom (over 25), complex environmental interactions, and the difficulty of identifying key parameters requiring control. While several studies \cite{aoi2022advanced,aoi2023maneuverable} have proposed different approaches, these efforts are often constrained by computational complexity, energy efficiency concerns, and limited adaptability to highly unstructured terrains.

\begin{figure*}[!h]
    \centering
    \includegraphics[width=18cm]{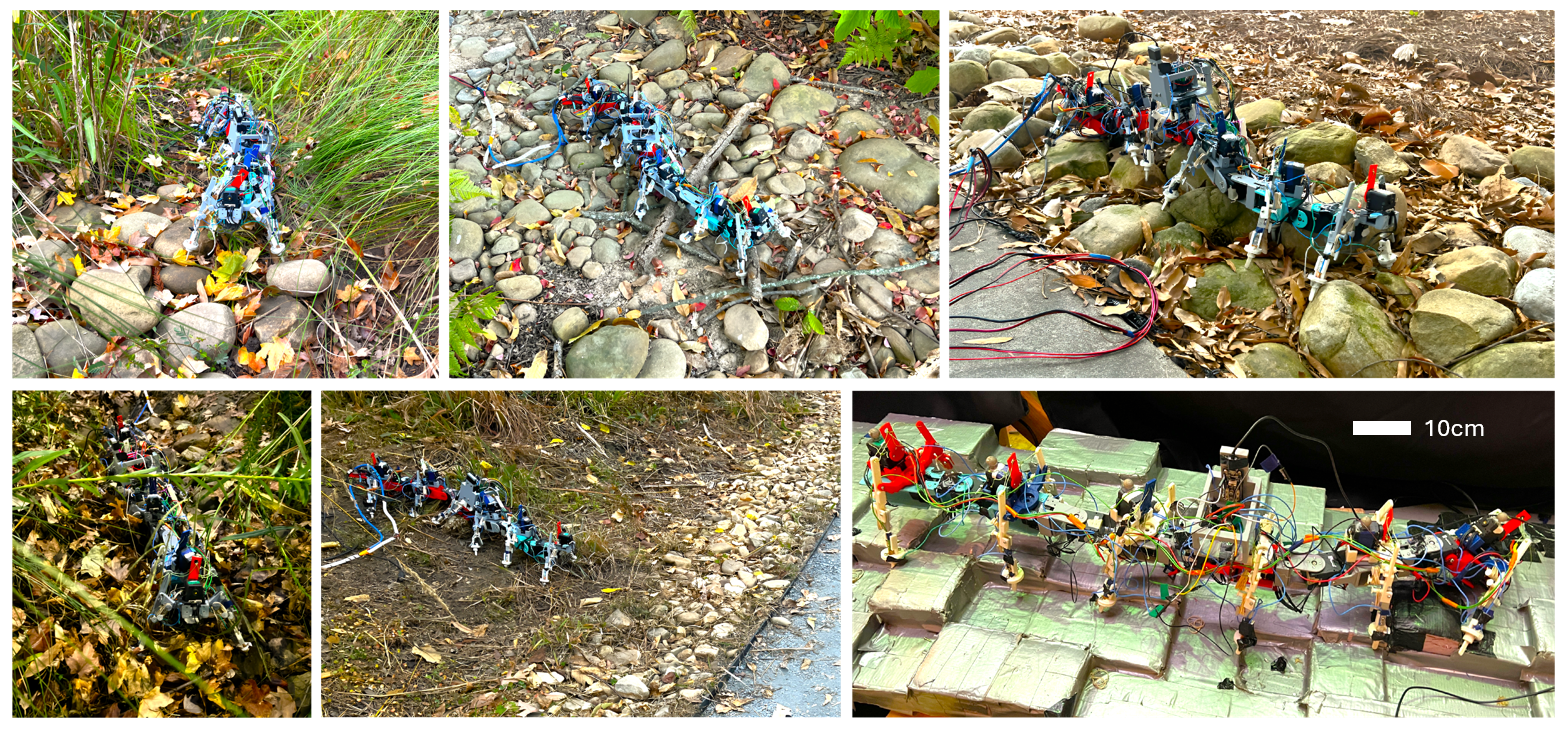}
    \caption{\textbf{Multi-legged robots navigating rugged landscapes.} To demonstrate the effectiveness of the proposed probabilistic model and feedback controller, we tested the robot in laboratory-based rough terrains (shown in the bottom right corner) and outdoor environments. Details of the lab-based terrain construction can be found in Section \ref{terrain construct}. The outdoor tests were conducted on rugged landscapes with a mixture of random tree debris, grass, boulders, mud, leaves, and rocks. }
    \label{fig:1}
\end{figure*}
Empirical evidence \cite{chong2023science} suggests that multi-legged robots can benefit from vertical body motion modulation when navigating rugged terrains with uneven surfaces and various obstacles. However, a notable gap exists in the literature regarding a systematic investigation of how vertical body undulation affects a robot's motion on rough terrain. Furthermore, the integration of vertical body undulation into a feedback control framework remains underexplored.

This paper systematically investigates the benefits of vertical body undulation in enhancing the speed of multi-legged robots using stochastic analysis, a widely adopted method for addressing uncertainties in robot-environment interactions\cite{prentice2009belief,majumdar2017funnel,mukadam2018continuous,yu2023stochastic,farshidian2017efficient,neunert2018whole,gangapurwala2020guided,osumi2006time}. Specifically, we develop two probabilistic models to analyze how the amplitude of vertical body waves affects the robot's speed on rough terrain and integrate this modulation into a feedback controller to improve robot's performance on rugged landscapes.

\begin{figure*}[!t]
    \centering
    \includegraphics[width=16.5cm]{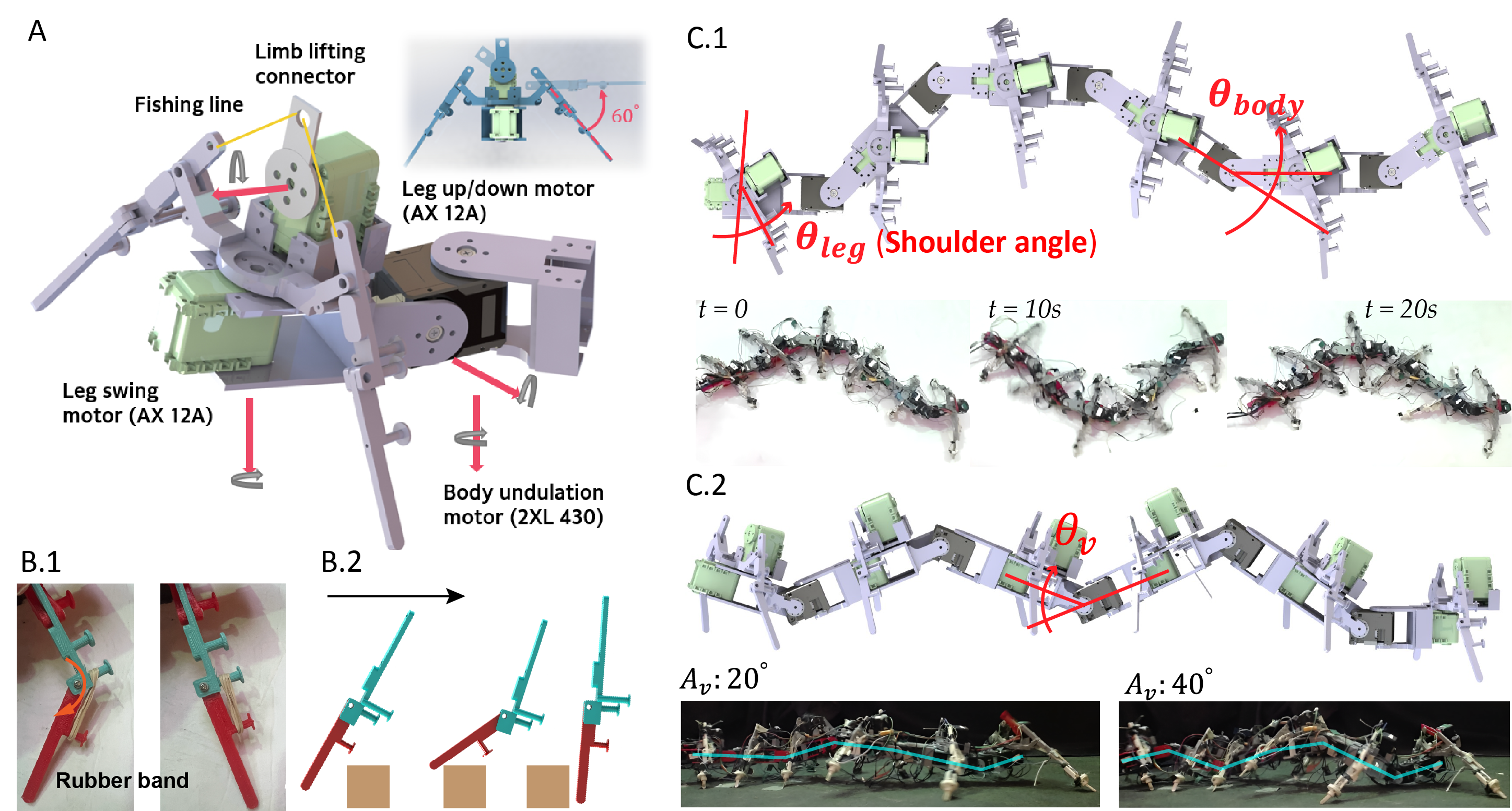}
    \caption{\textbf{Robophysical model description.} \textbf{A.} Design of a single module of the robophysical model, includes four active degrees of freedom: leg up and down, leg swing, and body lateral and vertical rotation. \textbf{B.} The compliant leg feature is achieved by a connected rubber band in 1, enabling the robot to overcome obstacles like those in 2. \textbf{C.} (1) Overhead view of the robotic system. Snapshots of the forward gait illustrate the coordination between horizontal body undulation and leg stepping. (2) Side view of the robotic system. Snapshots depicts variations in the vertical body undulation wave with different amplitudes ($A_v$).}
    \label{fig:rob_design}
\end{figure*}

To summarize our contribution, we first introduce a robotic system capable of generating coordinated body undulation and leg-stepping waves to enable forward motion. Next, we develop a binary contact sensing system to detect foot-ground contact. We then propose two probabilistic models to quantify the benefits of vertical wave modulation for multi-legged locomotion on rugged terrains. The first model predicts the robot's forward speed based on the actual-to-ideal foot-ground contact ratio, while the second model estimates this contact ratio given the terrain distribution and vertical wave amplitude. By combining these models, we demonstrate that: 1) a higher contact ratio generally leads to increased speeds, and 2) appropriately adjusting the vertical wave amplitude can effectively achieve this higher contact ratio. Utilizing these models, we design a feedback controller that allows the robot to optimize its vertical amplitude based on terrain roughness, estimated in real-time through contact ratio measurements. While our models and experiments primarily focus on terrains with varying height differences — one of the key causes of contact loss — the proposed framework is designed to address a broader class of rugged landscapes where such vertical irregularities are prevalent. We discuss the generalizability and limitations of this approach in Section \ref{Discussion and limitation}. Finally, we validate the controller's effectiveness through laboratory and outdoor experiments.

\section{Robot hardware design}
In this section, we explore the key design elements of our robotic model. Our approach builds on recent advancements in centipede-inspired robotics \cite{chong2022general,ozkan2021self,ozkan2020systematic,chong2023pnas,chong2023science}. As shown in Fig. \ref{fig:rob_design}, our robophysical model combines coordinated body undulation with limb movement to achieve forward motion. We incorporate compliant legs into the multi-legged robotic system to enhance the robot's navigation on challenging terrain \cite{ozkan2020systematic}. This design philosophy aims to create a versatile platform adaptable to a wide range of environments.

Our robotic system's thrust primarily originates from foot-ground interaction, in contrast to body-driven propulsion, which is more typical in robots operating continuously in environments like fluids or granular media
\cite{hatton2010generating,maladen2009undulatory,marvi2012friction}. When a leg contacts the ground and slips backward, it generates Coulomb friction with a forward-directed force. 
Terrain heterogeneity can disrupt the planned foot-ground contact, potentially reducing the robot's forward speed. Therefore, tracking the contact state to mitigate perturbations with a controller is an effective approach to enhance the robot's body forward velocity.

To further improve the robot's capabilities with feedback control, binary contact sensors are installed on each leg, as depicted in Figure. \ref{fig:sensor}. These sensors play a significant role in detecting foot-ground contact states, providing feedback signals crucial for the system's operation.

\subsection{Robophysical Model}
\label{robo_model}
Our robophysical model features a modular design with repeating segments, each containing three motors that control pitch and yaw of the body and leg assembly, resulting in four active degrees of freedom (DoF) per module (Fig. \ref{fig:rob_design}). Each module has two legs on opposing sides, with the front and back serving as connection points for subsequent modules. All modules components and connectors were 3D printed from PolyLite Polylactic Acid (PLA).

We incorporated compliant legs inspired by the passive morphology of centipedes observed during their interactions with its surroundings \cite{ozkan2020systematic}, enhancing the robots' ability to traverse rough terrain. Each leg consists of a two-bar linkage with a unidirectional hinge (knee joint) that bends opposite to the direction of forward motion. An elastic rubber band (Alliance Rubber 26324 Advantage Rubber Bands Size 32) restores the linkage to a neutral position after the leg encounters obstacles, aligning the links. This design ensures sufficient thrust during retraction, which is crucial for navigating uneven surfaces. The directional compliance facilitates a more evenly distributed contact area \cite{spagna2007distributed}, improving the robot's ability to traverse obstacles without significantly disrupting its gait. Experiments comparing the performance of rigid and compliant legs on rough terrain can be found in the Appendix \ref{rigid_vs_complaint}.

Each leg connects at a pivot joint, allowing it to pitch up to a maximum angle of $60^{\circ}$ (7 cm above the ground). A motor (AX-12A, DYNAMIXEL) cable-driven system alternates lifting opposing legs: rotating counterclockwise raises the left leg while the right remains grounded, and rotating clockwise raises the right leg while the left remains grounded. When cable tension decreases, another rubber band restores the leg to its original pitch angle. A second motor (AX-12A, DYNAMIXEL) controls the yaw ($\theta_{leg}$) of the leg assembly, facilitating leg swing. The yaw angle ($\theta_{leg}$) is defined as the shoulder angle for corresponding segment. Lastly, a 2-DoF motor (2XL430-W250, DYNAMIXEL) controls the yaw ($\theta_{body}$, Fig. \ref{fig:rob_design}.C.1) and pitch ($\theta_{v}$, Fig. \ref{fig:rob_design}.C.2) of the module, and connecting multiple modules facilitates lateral and vertical undulation (Fig.\ref{fig:rob_design}.C.1 and C.2). 

Through the assembly of multiple units, our robophysical model achieves locomotion via leg stepping, lateral, and vertical body undulation. Inspired by centipedes \cite{chong2023science,chong2023pnas}, the synchronized motion of leg stepping and lateral body undulation generates propulsive force necessary for forward movement (see Fig. \ref{fig:rob_design}.C.1). Additionally, the control over the vertical wave, as depicted in Fig. \ref{fig:rob_design}.C.2, provides valuable insights for future controller design endeavors.

\subsection{Binary contact sensing system design}

\begin{figure*}[!h]
    \centering
    \includegraphics[width=17cm]{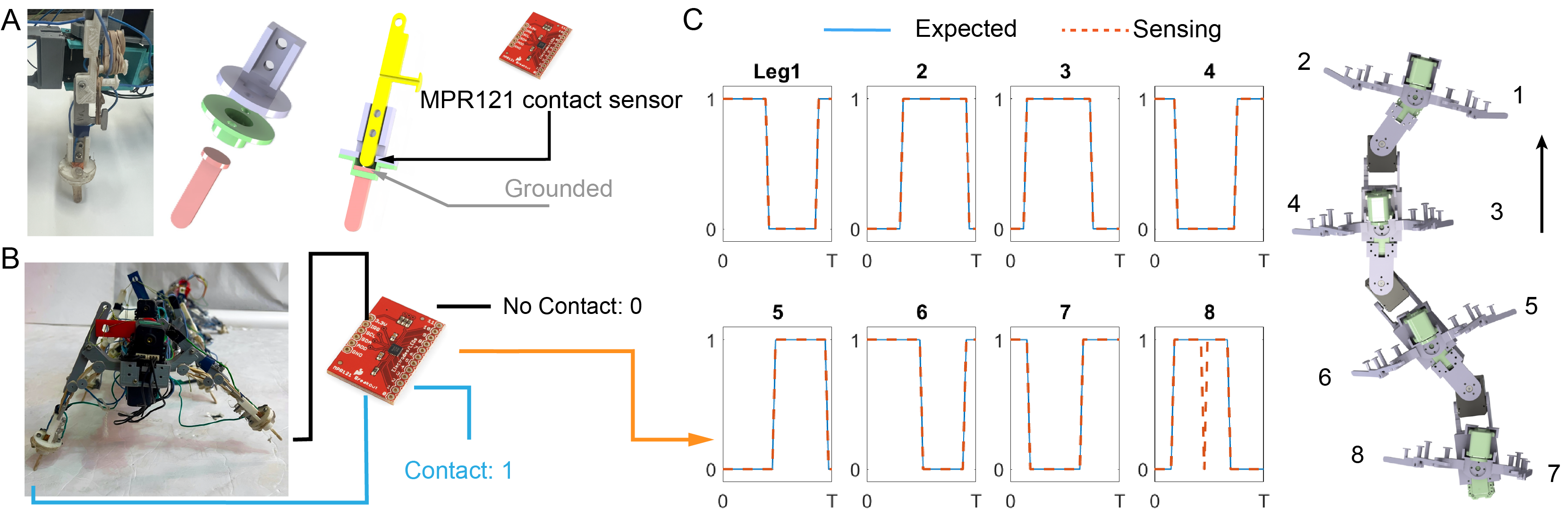}
    \caption{\textbf{Binary contact foot sensor system.} \textbf{A.} Design of a binary contact sensor for each foot, based on capacitive sensing. \textbf{B.} Contact state of the leg: 0 indicates no contact, while 1 indicates contact. \textbf{C.} Experimental and predicted foot-ground contact states of an 8-legged robot during one cycle of forward gait.}
    \label{fig:sensor}
\end{figure*}

We implemented a low-bandwidth binary contact sensor system (Fig. \ref{fig:sensor}.A) to monitor foot-ground interaction for each leg, allowing us to assess terrain heterogeneity and using this information as a feedback into our control system. Contact capacitive sensors (MPR121) embedded at the tip (highlighted in yellow in Fig.\ref{fig:sensor}.A) of each leg detect capacitance variance. The toe((highlighted in pink in Fig.\ref{fig:sensor}.A)) has a slight range of linear motion, resulting in minimal capacitance when the leg is suspended and maximal capacitance when it is grounded. 
The analog value shows a significant difference between the suspended state (greater than 200) and the grounded state (less than 5). Therefore, we classify any analog value below 50 as indicating contact. Figure.\ref{fig:sensor}.C illustrates experimental and predicted  contact states during the forward gait on level ground for an 8-leg robot, demonstrating strong alignment between the two.

\section{Robot gait design}
Building on the previously developed gait framework \cite{chong2023science,chong2023pnas,chong2022general} for generating forward motion through coordinated leg movement and horizontal body undulation, we enhance multi-legged locomotion across intricate landscapes by incorporating vertical body undulation. Empirical results \cite{chong2023science} demonstrate that integrating vertical body undulation in multi-legged robots with more than six legs plays a central role in feedback control strategies \cite{chong2023science}, enabling effective adaptation to varying terrains and mitigating environmental disturbances that impact the robot’s speed.

\subsection{Leg and body wave coordination} \label{leg and body wave}
We employ a binary variable $c$ to represent the leg's contact state, with $c=1$ denotes the stance phase and $c=0$ denotes the swing phase. In accordance with the methodology presented in~\cite{chong2022general}, the contact pattern for a multi-legged robot can be expressed as follows:

\begin{align}
     c_l(\tau_c,1) &=\begin{cases}
      1, & \text{if}\ \text{mod}(\tau_c,2\pi) < 2\pi D\\
      0, & \text{otherwise}
    \end{cases} \nonumber \\
    c_l(\tau_c,i) &= c_l(\tau_c - 2\pi\frac{\xi}{n}(i-1),1) \nonumber \\
    c_r(\tau_c,i) &= c_l(\tau_c+\pi,i),   
    \label{eq:contact}
\end{align}

\noindent where $\xi$ represents the number of spatial waves on legs, $D$ the duty factor, and $c_l(\tau_c,i)$ ($c_r(\tau_c,i)$) denotes the contact state of the $i$-th leg on the left (right) at gait phase $\tau_c$, with $i\in \{1, ... n\}$ for a $2n$-legged systems (See Fig. \ref{fig:rob_design}). 

Legs generate self-propulsion by retracting from anterior to posterior during the stance phase to engage with the environment and protracting from posterior to anterior during the swing phase to disengage. Considering this, we employ a piecewise sinusoidal function to define the anterior/posterior excursion angles ($\theta_{leg}$) for a given contact phase ($\tau_c$), as previously defined:
\begin{align}
        \theta_{leg,l}(\tau_c,1)  &=\begin{cases}
      \Theta_{leg}\cos{(\frac{\tau_c}{2D})}, & \text{if}\ \text{mod}(\tau_c,2\pi)  < 2\pi D\\
      -\Theta_{leg}\cos{(\frac{\tau_c-2\pi D}{2(1-D)})}, & \text{otherwise},
    \end{cases} \nonumber \\
    \theta_{leg,l}(\tau_c, i) &= \theta_l(\tau_c - 2\pi\frac{\xi}{n}(i-1), 1) \nonumber \\
    \theta_{leg,r}(\tau_c, i) &= \theta_l(\tau_c + \pi, i) 
    \label{eq:legmove}
\end{align}

\noindent where $\Theta_{leg}$ represents the amplitude of the shoulder angle, while $\theta_{leg,l}(\tau_c,i)$ and $\theta_{leg,r}(\tau_c,i)$ denote the shoulder angles of the $i$-th left and right legs, respectively, at the contact phase $\tau_c$. The shoulder angle reaches its maximum ($\theta_{leg}=\Theta_{leg}$) at the swing-to-stance phase transition and its minimum ($\theta_{leg}=-\Theta_{leg}$) at the stance-to-swing phase transition. Additionally, we set $D=0.5$ \cite{chong2022general,chong2023science,chong2023pnas}, unless stated otherwise.

We then introduce lateral body undulation by propagating a wave from head to tail:
\begin{align}
    \theta_{body}(\tau_b,i)=\Theta_{body} \text{cos}(\tau_b - 2\pi\frac{\xi^b}{n}(i-1)),
    \label{eq:body_horizontal}
\end{align}

\noindent where $\theta_{body}(\tau_b,i)$ represents the angle of the $i$-th body joint at phase $\tau_b$, and $\xi^b$ indicates the number of spatial waves on the body. For simplicity, we assume equal spatial waves in both body undulation and leg movement, denoted as $\xi^b = \xi$, allowing us to prescribe lateral body undulation based on its phase $\tau_b$.

Consequently, the gaits of multi-legged locomotors, achieved through the superposition of body wave and leg waves, are characterized by the phases of contact, \textcolor{black}{$\tau_c$}, and lateral body undulation, $\tau_b$. As outlined in~\cite{chong2022general}, the optimal body-leg coordination, which phases body undulation to facilitate leg retraction, is given by \textcolor{black}{$\tau_c$}$=\tau_b-(\xi/N+1/2)\pi$. We fixed both $\Theta_{body}$ and $\Theta_{leg}$ at $30^{\circ}$ for all the experiments, as this coordination effectively mitigate environmental disturbances that impacted the robot's speed \cite{chong2023science}. 
\begin{figure*}[!t]
    \centering
    \includegraphics[width=17cm]{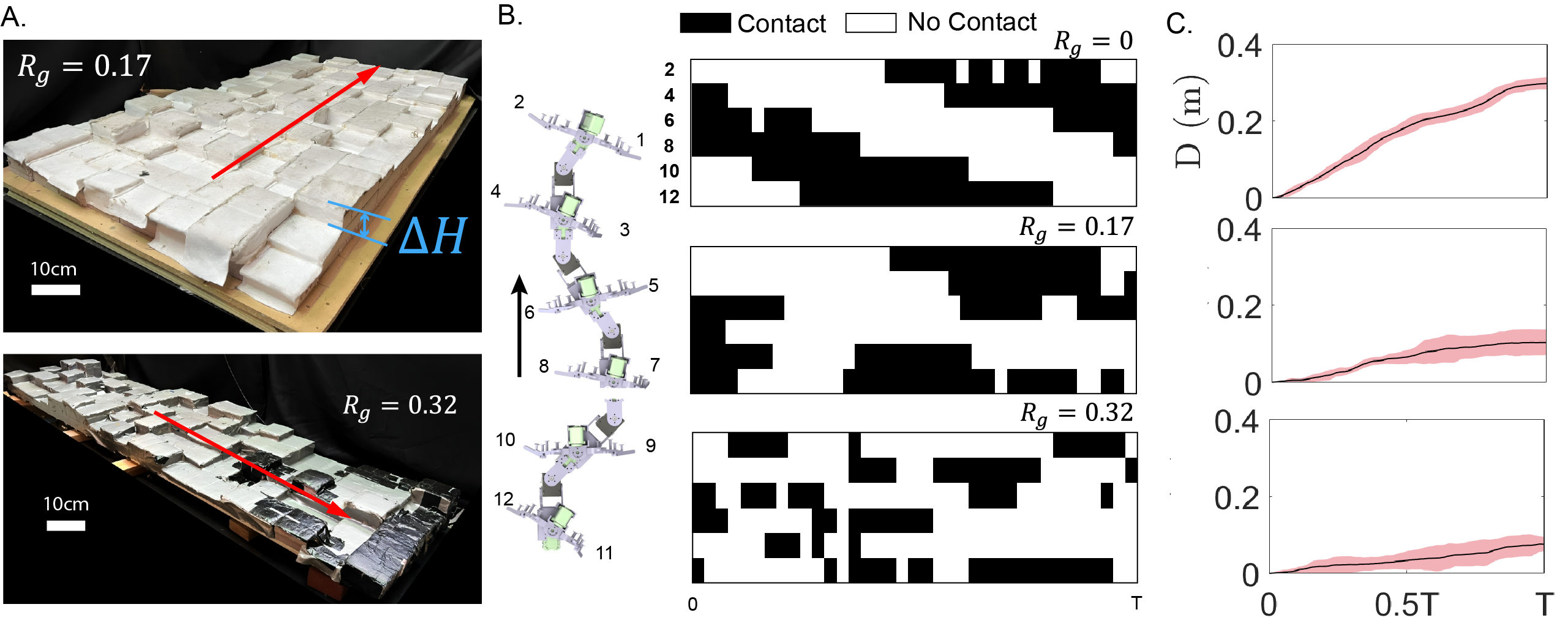}
    \caption{\textbf{Laboratory models of rugose terrain to study how rugosity affects locomotion dynamics and performance of the robot.} \textbf{A.} Two laboratory-based rugose terrains with varying levels of rugosity. Here, $R_g$ denotes the level of rugosity, while $\triangle H$ represents the height difference between two adjacent blocks along the longitudinal direction. The red arrow indicates the direction of the robot's movement during testing on these terrains. \textbf{B.} Depicting the contact states of the left half of a 12-legged robot on terrain with varying rugosity. A comparison between the flat terrain ($R_g=0$) and rough terrain ($R_g>0$) highlights examples of contact contamination (foot-ground contact state deviate from expected patterns due to terrain heterogeneity). \textbf{C.} The forward displacement of the 12-legged robot is measured across terrains of varying rugosity, showing the mean and standard deviation from 10 trials in each plot. Here, D represents the forward travel distance, while T represents one gait cycle, corresponding to 6 seconds in real time.}
    \label{fig:4}
\end{figure*}
\subsection{Vertical body undulation wave}
Here, we introduce vertical body undulation by propagating a wave along the body from head to tail:
\begin{align}
    \theta_v(\tau_b,i)=A_v \text{cos}(2\tau_b - 4\pi\frac{\xi^b}{n}(i-1)),
    \label{eq:body_horizontal}
\end{align}
\noindent where $\theta_v(\tau_b,i)$ represents the vertical angle of the $i$-th body joint at phase $\tau_b$ with an amplitude $A_v$. The vertical wave has a spatial frequency twice that of the lateral wave. This choice enables each body segment to oscillate vertically during the retraction phase, aiding the robot maintain foot-ground contact on rough terrain with varying elevations. More details could be found in Section \ref{sec vertical wave}. 

\section{Correlation Between Robot Speed and Foot-Ground Contact State} \label{speed correlation}

In a dissipation-dominated multi-legged system, thrust generation heavily depends on foot-ground contact states \cite{chong2023pnas,chong2023science}. In this section, we introduce a parameter, $\gamma$, which represents the fraction of robot contact with the ground during the retraction phase on rough terrain and characterizes the similarity between actual and flat ground contact states. Higher values of $\gamma$ indicate greater similarity to the flat ground contact state, suggesting improved locomotion efficiency and higher robot speed. However, no quantitative model currently exists to describe this correlation. To address this gap, we propose a probabilistic model to quantify the relationship between robot speed and $\gamma$.

\subsection{Rugose terrain construction} \label{terrain construct}
We constructed two terrains (Fig.\ref{fig:4}.A) with different levels of rugosity for the lab-based experiments. Each terrain is made of multiple 10 cm $\times$ 10 cm blocks with different heights. The height difference $\triangle H$ between adjacent blocks follows a normal distribution, denoted as $\triangle H \sim \mathcal{N} (0,\sigma(R_g))$, where the standard deviation of the height difference, $\sigma(R_g)$, is determined by the terrain rugosity $R_g$ and is calculated as:

\begin{equation}
\sigma = 15R_g \text{ (cm)}.
\end{equation}

For the $R_g = 0.17$ terrain (Fig.\ref{fig:4}.A), the terrain dimensions are (W, H) = (80, 160) cm, while for the $R_g = 0.32$ terrain, (W, H) = (50, 300) cm. More details regarding terrain construction can be found in the Supplementary Information (SI) of \cite{chong2023science}.
\begin{figure*}[!t]
    \centering
    \includegraphics[width=17cm]{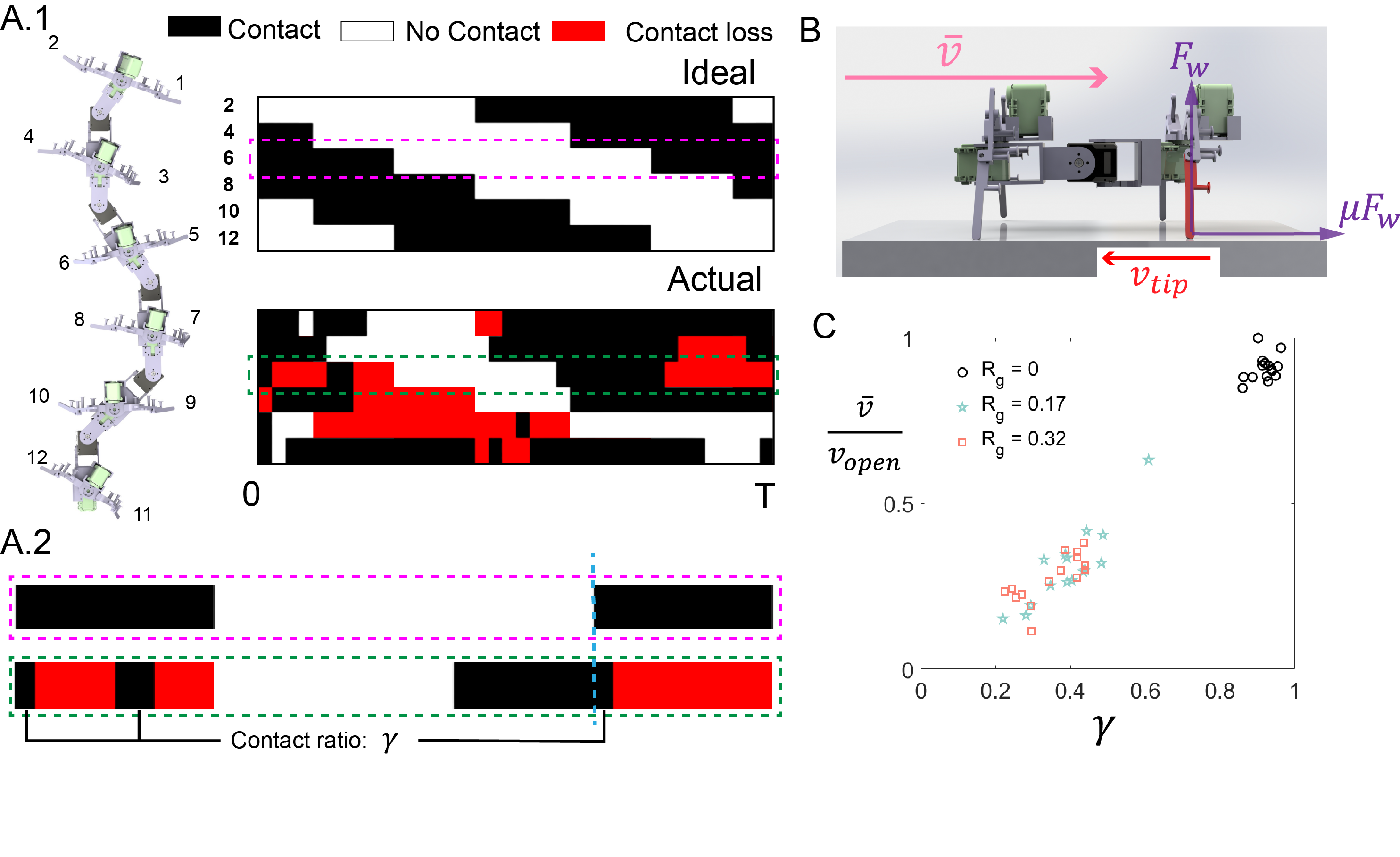}
    \caption{\textbf{Definition of the contact ratio $\gamma$ and experimental data  correlating speed with $\gamma$ in the absence of vertical undulation.} \textbf{A.} (1) Comparison between the ideal and actual foot-ground contact states is illustrated. In the ideal contact map, transitions from contact to no contact is categorized as "contact loss" and highlighted in red in the actual contact map. (2) The contact ratio, $\gamma$, represents the likelihood of the ideal and actual contact states matching, calculated as the average proportion of contacts that remain unchanged (areas not transitioning from black to red). \textbf{B.} A sketch depicting thrust generation by each leg of the robot. During the retraction or stance phase, the robot's leg contacts the ground and moves backward ($v_{\text{slip}}$), generating Coulomb friction that contributes to the forward velocity ($\bar{v}$). \textbf{C.} Experimental data reveal a correlation between the robot's speed and its contact ratio $\gamma$. Here, $\bar{v}$ represents the robot's average forward speed over a gait cycle, and $v_{\text{open}}$ denotes its forward speed in open space or on flat ground.}
    \label{fig:5}
\end{figure*}
\subsection{Contact state contamination }
When locomoting on rough terrain, irregularities deviate the robot's foot-ground contact state from the flat contact state, causing what we call contact state contamination. Fig. \ref{fig:4}.B illustrates examples of contact states contaminated affected by terrain rugosity. Since thrust generation in our robotic system heavily depends on the foot-ground contact state, contact state contamination may negatively impact the robot's speed. Fig. \ref{fig:4}.C illustrates how the robot's forward speed varies with different levels of contact state contamination.
\begin{figure*}[!t]
    \centering
    \includegraphics[width=17cm]{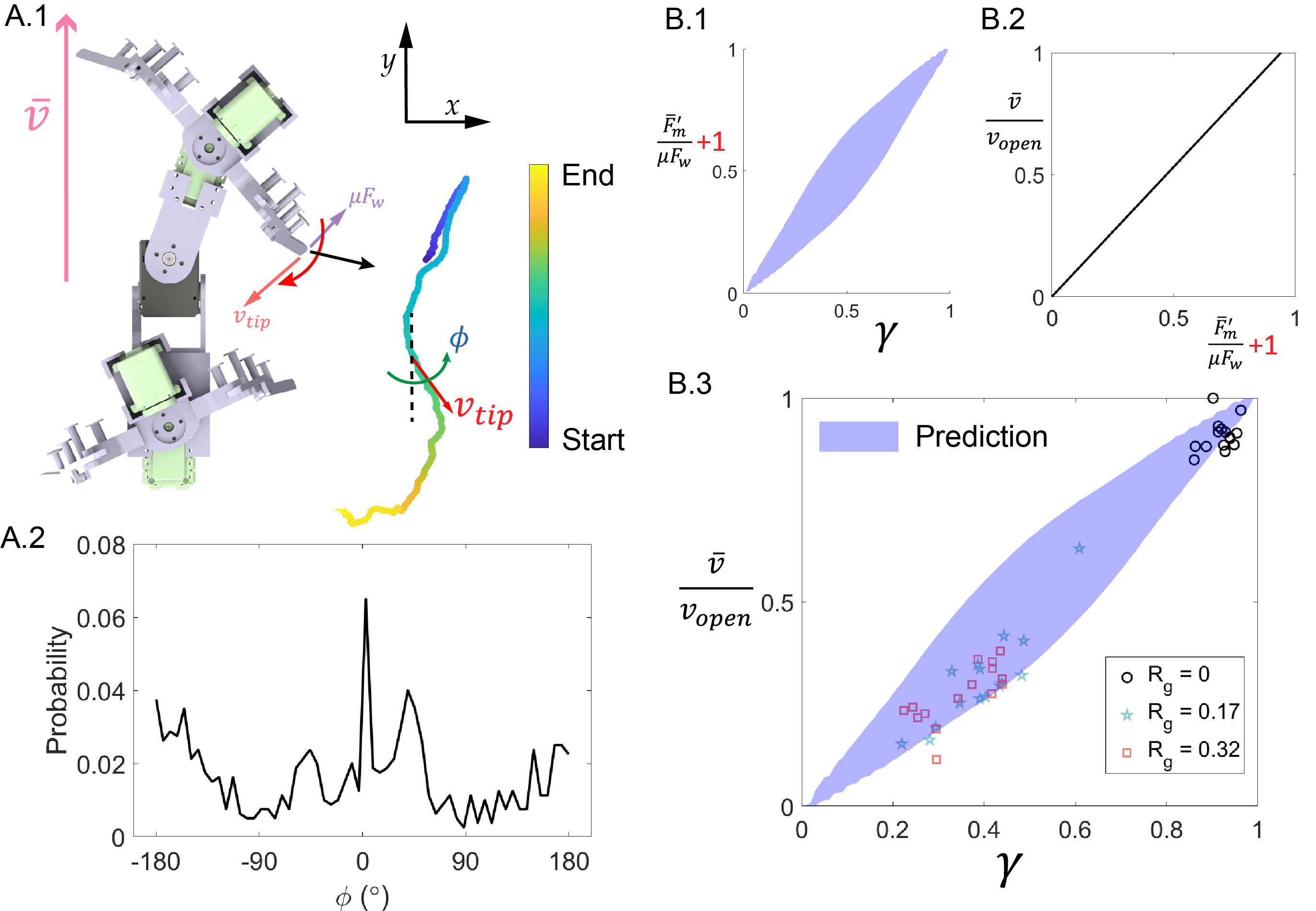}
    \caption{\textbf{Probabilistic model for correlation between robot speed and $\gamma$.} \textbf{A.} (1) The slipping trajectory of a leg during the retraction period is depicted. The slipping angle $\phi$ is defined as the angle between the robot's forward speed direction and leg's slipping direction. (2) Illustrated is the probability distribution of the slipping angle $\phi$ along the leg's slipping trajectory. \textbf{B.}  (1) and (2) show the correlation between the actual average friction $\bar{F}_m^{'}$ and the contact ratio $\gamma$, as well as the correlation between forward speed $\bar{v}$ and actual average friction $\bar{F}_m^{'}$. (3) The prediction (purple shaded area) of the probabilistic model for the speed-$\gamma$ correlation is shown, along with experimental data (scatter plots) for validation.}
    \label{fig:6}
\end{figure*}
\subsection{Contact ratio $\gamma$} 

Section \ref{leg and body wave} suggests that legs generate self-propulsion by retracting during the stance phase to establish contact with the environment and protracting during the swing phase to disengage. Thus, the proportion of foot-ground contact retained during the retraction period directly influences robot's final thrust. 

To quantify the likelihood between the actual and ideal contact state, we introduce a parameter $\gamma$, the effective contact ratio. We define the duration of the retraction period as $T_p$, corresponding to the duration of the black region in the ideal contact state map shown in Figure \ref{fig:5}.A.1. We then discretize this period into $K$ independent time steps, dented as $t_i$. At each time step, $c_a(i)$ represents the actual contact state, where $c_a(i) = 1$ indicates contact, and $c_a(i) = 0$ signifies no contact. Thus, mathematically, the contact ratio $\gamma$ can be expressed as:5
\begin{equation}
    \gamma = \frac{\sum_{i=1}^K c_a(t_i)}{K}
\end{equation}

As depicted in Fig. \ref{fig:5}.A, contact loss is defined as the transition from contact to no contact compared to the ideal contact state map. The contact ratio is then computed as the average proportion of contact maintained during the retraction period across all legs over a cycle of motion. We disregard the impact of contact during the protraction period on reducing thrust, as the flexible leg can minimize contact forces during this phase. Specifically, when the leg makes contact with the ground during the protraction period, it undergoes deformation as it moves from the posterior to the anterior end. This deformation causes the supporting force to decrease, as its origin shifts from supporting body weight to the torque from a soft rubber band.

\subsection{Probabilistic model for correlation between robot speed and $\gamma$}
Thus far, we have defined a parameter $\gamma$ that directly influences the thrust generation of the robot. As shown in Fig. \ref{fig:5}.B, during the retraction period or stance phase, the robot's leg makes contact with the ground and swings backward (\textcolor{black}{$v_{\text{tip}}$}), resulting in Coulomb friction components in the direction of forward speed ($\bar{v}$). The contact loss could decrease $\gamma$ and potentially reduce the forward speed of the robot. The subsequent objective is to elucidate the relationship between the velocity of the robot and $\gamma$. Our experimental results in Fig. \ref{fig:5}.C demonstrate a positive correlation between the robot's speed and $\gamma$. Next, we  derive a probabilistic model to describe the correlation between the robot's speed $\bar{v}$ and $\gamma$ in sequential order, 
\begin{itemize}
    \item Correlation between friction and $\gamma$.
    \item Correlation between friction and $\Bar{v}$.
    \item Correlation between $\Bar{v}$ and $\gamma$. 
\end{itemize}

\subsubsection{Correlation between friction and $\gamma$} \label{thrust_gamma}
Fig.\ref{fig:6}. A.1 shows the typical trajectory of a foot of a 12-legged robot during retraction period or stance phase. We quantify the slipping direction, denoted by $\phi$, as the angle between the direction of motion of a tip and the forward movement direction of the robot. We discretize the continuous slipping trajectory into K discrete points. Thus, the average friction of the m-th leg over a cycle of motion can be computed as,
\begin{equation}
    \Bar{F}_m = \frac{\sum_{i=1}^{K}\mu F_w \cos(\phi_i)}{K} 
\end{equation}
where $\mu$ is the friction coefficient and $F_w$ (Fig. \ref{fig:5}.B) denotes the average ground supporting force on the foot when the robot's weight is evenly distributed across each landing foot. It is noteworthy that $\bar{F}_m=0$ signifies the robot reaching a steady state, where the thrust and drag nullify each other over a cycle of motion. \cite{maladen2009undulatory,maladen2011mechanical,maladen2011undulatory}

Figure \ref{fig:6}.A.2 shows the probability distribution of the slipping angle $\phi$ for the first leg (Figure \ref{fig:4}) of a 12-legged robot. To derive this distribution, we conducted three cycles of motion with the robot on flat ground, each repeated for 10 trials. The trajectory of the leg was meticulously tracked using our motion tracking system (OptiTrack; see Appendix). By computing the tangential direction of the nearest point on the trajectories, we obtained the distribution of slipping angles.

Note that the probability distribution of $\phi$ is invariant to the choice of leg because each leg repeats the cyclical motion with a phase lag according to Section \ref{leg and body wave}. We represent the probability for a specific slipping angle $\beta_s \in [-180^{\circ},180^{\circ}]$ as $Pr(\beta)=Pr(\phi=\beta)$. Thus, we can formulate $\gamma$ as follows:

\begin{equation}
\gamma = \sum_{i=1}^{n} Pr(\beta_i) \cdot w_i,
\end{equation}
where $w_i \in [0,1]$ denotes the proportion of contact remaining undisturbed by terrain rugosity for a specific slipping angle $\beta_i$. For instance, $w_i$ equals 1 on an ideal terrain. However, when the robot traverses rough terrain, the foot may lose some contact proportion, leading to a probable decrease in $w_i$. $n$ denotes the number of bins used to partition the probability distribution. Given a contact ratio $\gamma$ and the empirical distribution of $Pr(\beta)$, we did perform numerical search to obtain all possible numerical solution for $w_i$. 

Similarly, the average friction $\Bar{F}$ can be rewritten in a probabilistic way,
\begin{equation}
    \Bar{F}_m = \sum_{i=1}^{n} \mu F_w \cdot \cos(\phi=\beta_i) Pr(\beta_i).
    \label{ friction prob}
\end{equation}

To simplify our calculation, we assume that when contact loss occurs, the support force exerted by the affected landing leg is transferred to the belly of the robot and acts as a frictional drag against the robot's moving direction. The actual averaged friction for the $m$-th leg is then computed as, 
\begin{equation}
    \Bar{F}^{'}_m = \sum_{i=1}^{n} \mu F_w w_i \cdot \cos(\phi=\beta_i) Pr(\beta_i)-\mu F_w (1-w_i).
\end{equation}

As defined previously, $w_i$ represents a weighting factor indicating the extent of disturbance on a particular slipping direction, with $w_i \leq 1$. If $w_i = 1$ for all $i \in [1,n]$, where $n$ is the number of slipping direction bins, then $\bar{F}_m = \bar{F}^{'}_m$. If $w_i < 1$, the force disturbance can be computed as $\delta \bar{F}= \bar{F}_m - \bar{F}^{'}_m$. Given that $\bar{F}_m=0$, it follows that $\delta \bar{F} = -\bar{F}^{'}_m$. By tracking the variation of  $\bar{F}^{'}_m$ due to contact state contamination, we could compute how the force disturbance ($\delta \bar{F}$) changes with disturbances in the contact state.

Since $\Bar{F}^{'}_m$ is determined by $w_i$ and $w_i$ can be determined by $\gamma$, we establish the relationship between $\Bar{F}^{'}_m$ and $\gamma$. Here, we normalize $\Bar{F}^{'}_m$ by $\mu F_w$, denoted as $\frac{\Bar{F}^{'}_m}{\mu F_w}$. This correlation is illustrated in Fig. \ref{fig:6}.B.1.

A recent study by Chong et al. \cite{chong2023pnas} established a  linear relationship between the actual speed of a robot and its disturbed averaged friction ($\delta \bar{F}$). Leveraging these findings, we now investigate the correlation between velocity and friction within the context of our specific scenarios.

\subsubsection{Correlation between friction and $\Bar{v}$}
The models presented in \cite{chong2023pnas,chong2023science} offer insights into predicting a robot's speed $\bar{v}$ based on its averaged friction $\bar{F}$. Specifically, they propose an effective force-velocity relationship expressed as follows:

\begin{equation}
\delta\bar{F} = -C \delta{\bar{v}}
\label{dF_dv}
\end{equation}

Here, $\delta\bar{F}$ represents the disturbance to the average friction, while $\delta{\bar{v}}$ $ = \bar{v}-v_{open}$ denotes the disturbance to the speed. The constant $C$ is determined by the robot's dimensions and its ideal steady-state speed $v_{open}$. Specifically, $C = \int^{\pi}_0 \mu F_w/v_{open} \sin(\tan ^{-1}(v_{open}/v_x(\tau))d\tau$. Here, $v_{x}(\tau)$ is the lateral speed of the leg tip over time and can be obtained by analyzing robot's geometry \cite{chong2023pnas}.

In the previous section, we discussed how disturbances to the contact state result in a change in friction. Specifically, we can observe $\delta\bar{F}$ by considering the ratio $\frac{\Bar{F}^{'}_m}{\mu F_w}$. Equation \ref{dF_dv} allows us to anticipate the relationship between friction and velocity. Figure \ref{fig:6}.B.2 illustrates our prediction of the correlation between $\frac{\Bar{F}^{'}_m}{\mu F_w}$ and $\frac{\bar{v}}{v_{open}}$ using this model. For the robotic system under consideration, we approximate this relationship as:
\begin{equation}
    \frac{\bar{v}}{v_{open}} \approx 1.065(\frac{\Bar{F}^{'}_m}{\mu F_w}+1)
    \label{v_Fm}
\end{equation}

\subsubsection{Correlation between $\Bar{v}$ and $\gamma$}
In the previous sections, we introduced a model to predict average friction based on the contact ratio $\gamma$. By analyzing the friction disturbance using the dimensionless ratio $\frac{\Bar{F}^{'}_m}{\mu F_w}$, we can calculate the force disturbance $\delta \bar{F}$ as a function of $\gamma$. Since Eq.\ref{v_Fm} establishes a linear relationship between robot's speed and the actual friction, this enables us to link the robot's speed to the contact ratio $\gamma$.

We assume that the linear force-velocity relationship in Eq.\ref{v_Fm} remains valid even on rough terrain. Experimental data presented in the following sections support this assumption. Consequently, we can determine the relationship between $\bar{v}$ and $\gamma$ by integrating the correlations derived in the previous subsections. Figure \ref{fig:6}.B.3 illustrates our predicted correlations between $\bar{v}$ and $\gamma$, alongside experimental results for comparison.

\section{Vertical body motion mitigates environmental disturbance} \label{sec vertical wave}
In Section \ref{speed correlation}, we demonstrated that the robot's velocity is approximately proportional to the effective contact ratio $\gamma$. Therefore, mitigating environmental disturbance on $\gamma$ effectively reduces disturbance on velocity.
\begin{figure*}[!t]
    \centering
    \includegraphics[width=17cm]{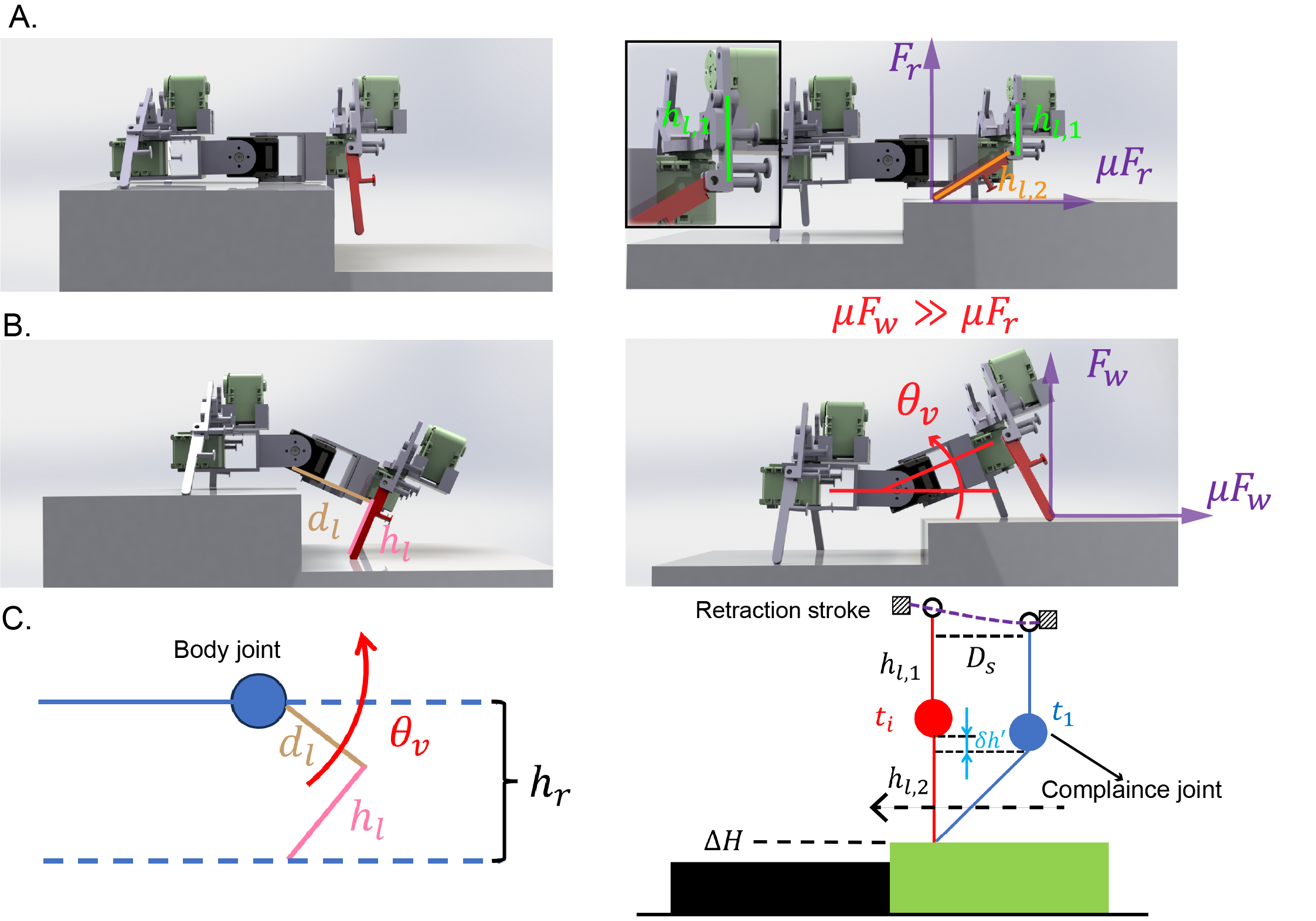}
    \caption{\textbf{Vertical body undulation serves as a mechanism to alleviate foot-ground contact state distortion.} \textbf{A. Examples of contact loss on rough terrain.} (left) Contact loss occurs when the leg's length is insufficient to reach the ground. (right) Contact loss occurs when the flexible leg is deformed by shifts in terrain level. In this scenario, the supporting force $F_r$ diminishes as it transitions from supporting the body weight to generating torque on the rubber band. \textbf{B. Vertical body undulation reduces the contact contamination.}  (left) By lowering the leg deeper, contact contamination is minimized. (right) Conversely, lifting the leg facilitates its return to its initial undeformed state, thereby reducing contact contamination. \textbf{C. Sketch for deriving probabilistic models:} An illustration depicting the geometric relationship between the robotic structure and variations in terrain height. (left) This sketch illustrates the maximum depth ($h_r$) that the leg can reach with the vertical joint pitch motion. (right) This sketch demonstrates how leg retraction restores the leg from deformation to its straight state. Given the retraction distance $D_s$, the compliance (knee) joint is lifted by $\delta h'$. Circles without fill on the dashed line represent the pivot joint for leg lifting, while solid circles indicate the knee joint for leg compliance. $h_{l,1}$ is distance between the pivot joint and knee joint while $h_{l,2}$ is the distance between the knee joint and leg tip. Squares represent the stroke for leg retraction, showing the range of leg swing from anterior to posterior during the stance phase of one motion cycle.}
    \label{fig:7}
\end{figure*}
As shown in Fig. \ref{fig:7}.A, contact loss can occur in two scenarios due to the terrain level shifts. First, when the leg length is insufficient to reach the ground. Second, when the leg  deforms, leading to a reduction in support force. In the latter case, the supporting force shifts from the body weights to the elastic force generated by the rubber band attached to each flexible leg joint. To quantify the terrain level shift, We define the height difference as:
\begin{equation}
    \triangle H = H(t^+) - H(t),
\end{equation}
where $H(t)$ and  $H(t^+)$ represent the terrain height at the current and next time steps, respectively.

We observe that implementing vertical motion in each body joint enhances the robot's ability to maintain contact on challenging and uneven terrain. As shown in Fig.\ref{fig:7}, the robot can recover its contact state by pitching the body segment downward, bringing the leg into contact with the ground, or pitching it upward to correct any deformation in the leg. Unlike the discrete foothold planning used in bipedal or quadrupedal robots, our approach models the vertical body undulation of a multi-legged robot using sinusoidal traveling waves. 

\subsection{Probabilistic model predicts $\gamma$ on rough terrain} \label{model vertical wave}
\subsubsection{Gait without vertical wave}
We begin our analysis by considering the simplest scenario: the robot's gait without vertical body undulation waves. To address the variations in height difference, we categorize them into two cases: $\triangle H > 0$ and $\triangle H \leq 0$. To dynamically control the sign of $\triangle H$, we introduce a switch denoted as $S$:

\begin{equation}
S = \begin{cases}
1 & \text{Pr: } p_1 \quad \quad \quad\text{for } \triangle H \leq 0 \\
2 & \text{Pr: } 1 - p_1 \quad \text{ for } \triangle H > 0.
\end{cases}
\label{switch}
\end{equation}

This switch enables us to control the sign of $\triangle H$ according to the specified probabilities.

If $\triangle H \leq 0$, it indicates that the terrain at the next time step is lower than the current one. Contact loss occurs when the robot leg is not long enough to touch the ground (see Fig. \ref{fig:7}.\textbf{A}). In probability terms,
\begin{equation}
    P_{loss,1} = Pr(\triangle H>h_l|S =1),
\end{equation}
where $h_l$ (Fig. \ref{fig:7}.B )is the maximum depth the robot can reach without utilizing any vertical body movement.

If $\triangle H > 0$, it indicates that the terrain at the next time step is higher than the current one. In this case, the situation is more complex because the leg can still make contact with the ground. Our robot's flexible legs can bend when encountering higher terrain. We assume a contact loss occurs if the robot leg deforms upon encountering higher terrain (see Fig. \ref{fig:7}.\textbf{B}, right). 

The contact is then restored by the leg's retraction movement, represented by the retraction distance $D_s(t)$. During the retraction process, we assume the leg remains fixed at the foot-ground contact point (see Fig. \ref{fig:7}.\textbf{C}, right).
Then, we discretize the retraction period over a motion cycle into $m$ time steps. For a given time step $t_i$ during the retraction period of the robot's leg, we define the maximum height difference that can be recovered by retraction as $\delta h(t_i)^{'}$, which is computed as:
\begin{equation}
    \delta h(t_i)^{'} = h_{l,2} (1-\cos(\arcsin(\frac{D_s(t_i)}{h_{l,2}})))
\end{equation}
The probability for contact loss could be expressed as:
\begin{equation}
    P_{loss,2} = \frac{\sum_{i=1}^{m} Pr(\triangle H> \delta h(t_i)^{'}|S=2)}{m}
\end{equation}
The cumulative probability of contact loss, $P_{loss}$, arising from two distinct sources can be expressed as the sum of the products of the probability of encountering each source ($p_1$ and $1-p_1$) and the corresponding probability of contact error ($P_{loss,1}$ and $P_{loss,2}$). This relationship is given by the equation:
\begin{equation}
    P_{loss} = p_1 P_{loss,1}+(1-p_1) P_{loss,2}
\end{equation}

\textcolor{black}{
Note that in Section~\ref{speed correlation}, Fig.~\ref{fig:5}A, we define \textit{contact loss} as regions where contact is lost—specifically, transitions from contact (on flat ground) to no contact (on rugged terrain)—relative to the ideal contact pattern.
In Section~\ref{sec vertical wave}, we extend the definition of contact loss to include both:
\begin{itemize}
    \item \textbf{Contact $\rightarrow$ No contact:} This results in a missed step, as shown in Fig.~\ref{fig:7} (left).
    \item \textbf{Contact $\rightarrow$ Contact, but with leg deformation:} Although the foot maintains contact with the ground, the leg fails to maintain its intended shape or posture. In the stance phase, where ground contact is planned, such deformation is considered a form of contact loss, as illustrated in Fig.~\ref{fig:7} (right).
\end{itemize}
}

\subsubsection{Gait with vertical wave}

We introduce an additional body undulation wave, the vertical wave, to mitigate contact loss. Intuitively, the vertical wave helps the leg reach deeper areas when $\triangle H \leq 0$, and it can lift the robot leg to avoid or reduce bending when $\triangle H > 0$. The amplitude of these vertical waves is defined as $A_v$.

If $\triangle H \leq 0$, the maximum depth that the leg can reach is defined as $h_r(t)$ (Fig.\ref{fig:7}.\textbf{C}, left). This depth, $h_r$, is computed as:
\begin{equation}
h_r(t) = d_l \sin(\theta_v(t))+ h_l \cos(\theta_v(t))    
\end{equation}

Thus, the new probability of contact loss, considering the vertical wave modulation, can be expressed as:
\begin{equation}
    P_{loss,1} = \frac{\sum_{i=1}^{m}Pr(\triangle H>h_r(t_i)|S =1)}{ m },
\end{equation}

If $\triangle H > 0$, the vertical wave introduces an offset to $\triangle H$ by lifting the robot leg (Fig. \ref{fig:7}.\textbf{C}, right). The new height difference with the offset can be expressed as:
\begin{equation}
    \triangle H^{'} = \triangle H + h(t),
\end{equation}
where $h(t)$ represents the offset controlled by the vertical wave amplitude $A_v$ and time $t$. The probability of contact error due to this offset can be expressed as:
\begin{equation}
    P_{loss,2} = \frac{\sum_{i=1}^{m} Pr(\triangle H^{'} >  \delta h(t_i)^{'}|S=2)}{m}.
\end{equation}

\subsubsection{Contact ratio $\gamma$ prediction}

\textcolor{black}{As a reminder, the contact ratio, denoted as $\gamma$, is defined as the likelihood between the actual contact state and the ideal contact state during the retraction period.} The computation of the contact ratio is expressed by the formula:
\begin{equation}
    \gamma = 1-P_{loss}.
\end{equation}

In Fig.\ref{fig:7}.A, we define contact error as contact flipped by terrain noise compared to the contact state on flat ground. Thus, we can calculate the probability of contact error:
\begin{equation}
    P_e = \frac{1-\gamma}{\gamma'},
\end{equation}
where $\gamma'$ is the contact ratio over one gait cycle on ideal terrain. $\gamma'$ is controlled by the vertical wave amplitude, $A_v$. As $A_v$ increases, $\gamma'$ decreases when locomoting on ideal terrain. This happens because the vertical body undulation can cause the leg to lift off the ground during the retraction period or stance phase, leading to a reduced contact ratio, with $\gamma'$ being less than 1.\\

\subsubsection{Experiment validation}
As defined in Section \ref{terrain construct}, we assume that the height difference $\triangle H$ follows a normal distribution $(0,\sigma)$, where the variance $\sigma$ is regulated by the terrain noise level $R_g$. Specifically, we define $\sigma = 15R_g $. Thus, we can write:
\begin{equation}
    \triangle H \sim \mathcal{N} (0,\sigma(R_g)).
\end{equation}
\begin{figure}[!t]
    \centering
    \includegraphics[width=8cm]{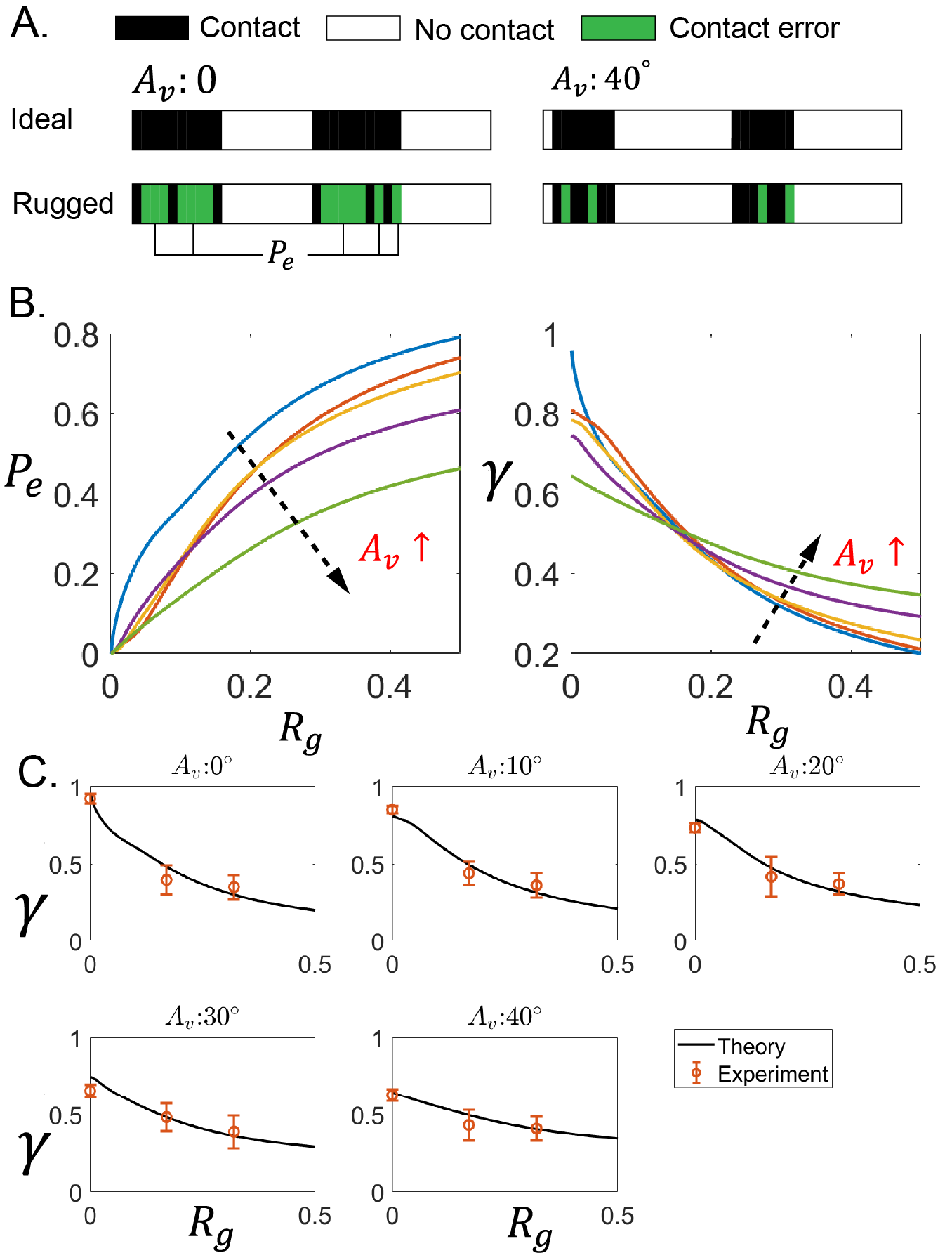}
    \caption{\textbf{Theoretical predictions and experimental validation of probabilistic models for the $\gamma-A_v$ correlation.} \textbf{A.} Definition of contact error: the probability of contact error, denoted as $P_e$, is the portion of contacts that deviate from their corresponding flat terrain contact states. \textbf{B.} Probabilistic predictions of $P_e$ and $\gamma$ for various vertical amplitudes ($A_v$) across terrains with different rugosity. \textbf{C. } Experimental validation of the probabilistic models. \textcolor{black}{We conducted 10 experimental trials for each parameters. The error bars represent the standard deviation.}}
    \label{fig:8}
\end{figure}
Note that $\triangle H$ exhibits an equal likelihood of being positive or negative when it follows a normal distribution. Consequently, $p_1$ in Eq.\ref{switch} equals 0.5. By substituting the dimensions of the robot into the probabilistic models, we obtain predictions for $P_e$ and $\gamma$ on terrains of different rugosity, as shown in Fig.\ref{fig:8}.B. To validate our model, we conducted laboratory  experiments using our 12-legged robot on terrains with rugosities $R_g =0, 0.17, 0.32$ and different vertical wave amplitudes. The experimental results, presented in Fig.\ref{fig:8}.C, closely match our theoretical predictions.

\subsection{Vertical motion reduces environmental disturbance}
Contact loss, shown in Fig. \ref{fig:7}.A, occurs due to terrain level shifts in two scenarios: insufficient leg length preventing ground contact and leg deformation. Vertical body undulation helps the robot restore contact by either increasing leg placement depth for ground contact or lifting the leg to correct deformation.
\begin{figure}[!h]
    \centering
    \includegraphics[width=8.5cm]{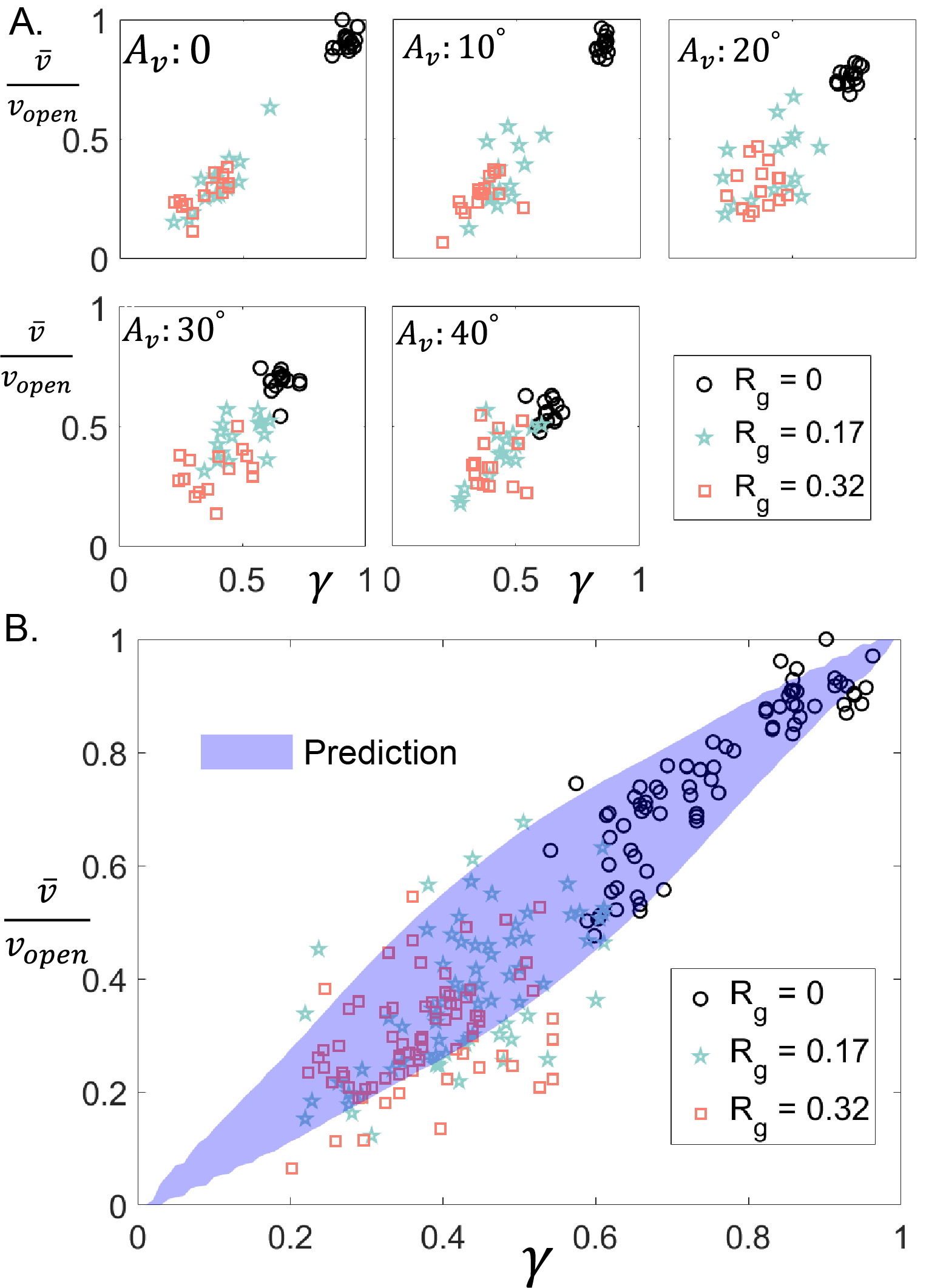}
    \caption{\textbf{Experimental data and probabilistic model predictions for speed-$\gamma$ correlations.} \textbf{A.} Experimental data were collected by testing the robot on terrains with varying rugosity while adjusting the vertical amplitude $A_v$. \textbf{B.}  The experimental results validate the accuracy of the probabilistic model in predicting the relationship between the robot's forward speed $\bar{v}$ and the contact ratio $\gamma$.}
    \label{fig:9}
\end{figure}
Similar to leg stepping and horizontal body undulation, we program vertical body undulation as a traveling sinusoidal wave, but with twice the spatial frequency of the horizontal wave. This design allows each robot segment to move both upward and downward during the retraction phase. Since rough terrain often exhibits alternating level shifts, encoding vertical body motion helps the robot address both types of contact loss caused by these terrain variations.

As defined in Section \ref{terrain construct}, terrain rugosity ($R_g$) is determined by the variance in height difference between two adjacent blocks. A higher vertical body undulation amplitude ($A_v$) enables the robot to maintain contact even when there are significant height differences between blocks. However, increasing $A_v$ reduces the contact ratio $\gamma$ on flat or less rough terrain, as the vertical body wave lifts the leg during the retraction phase. Intuitively, a higher $A_v$ increases  the probability of contact error ($P_e$) more gradually as terrain rugosity increases. Additionally, while higher $A_v$ causes the contact ratio $\gamma$ to start at a lower value, it decreases more slowly as terrain rugosity increases. Therefore, increasing the vertical amplitude can reduce environmental disturbances, improving both the robot's contact state and its speed.

We first verify this assumption using probabilistic model from Section \ref{model vertical wave}. As depicted in Fig. \ref{fig:8}.B, with a high vertical amplitude $A_v$, the probability of contact error $P_e$ increases more gradually as terrain rugosity increases, and the contact ratio $\gamma$ is less sensitive to terrain noise shifts.

To verify our hypothesis through more real-world experiments, we tested our 12-legged robot on terrain with varying rugosity using different vertical amplitudes $A_v$. A motion tracking system and binary contact sensors were used to monitor the robot's speed and contact ratio $\gamma$, respectively. The results (Fig. \ref{fig:9}.A) show as $A_v$ increases, both the robot's speed and contact ratio $\gamma$ tend to concentrate around the central area near (0.5, 0.5). \textcolor{black}{Specifically, on terrain with $R_g =0.17$, the optimal vertical amplitude $A_v$ increases the average contact ratio $\gamma$ by 25\% and speed by 30\%, compared to the gait without vertical wave ($A_v = 0$). On terrain with $R_g = 0.32$, the optimal $A_v$ increases the average $\gamma$ by 18\% and speed by $20\%$. } These findings suggest that vertical body undulation effectively helps the robot mitigate environmental disturbances, improving both its speed and contact state.

\section{Enhance Robot Performance on Rugged Landscapes Through Feedback Control}
\begin{figure}[!h]
    \centering
    \includegraphics[width=8cm]{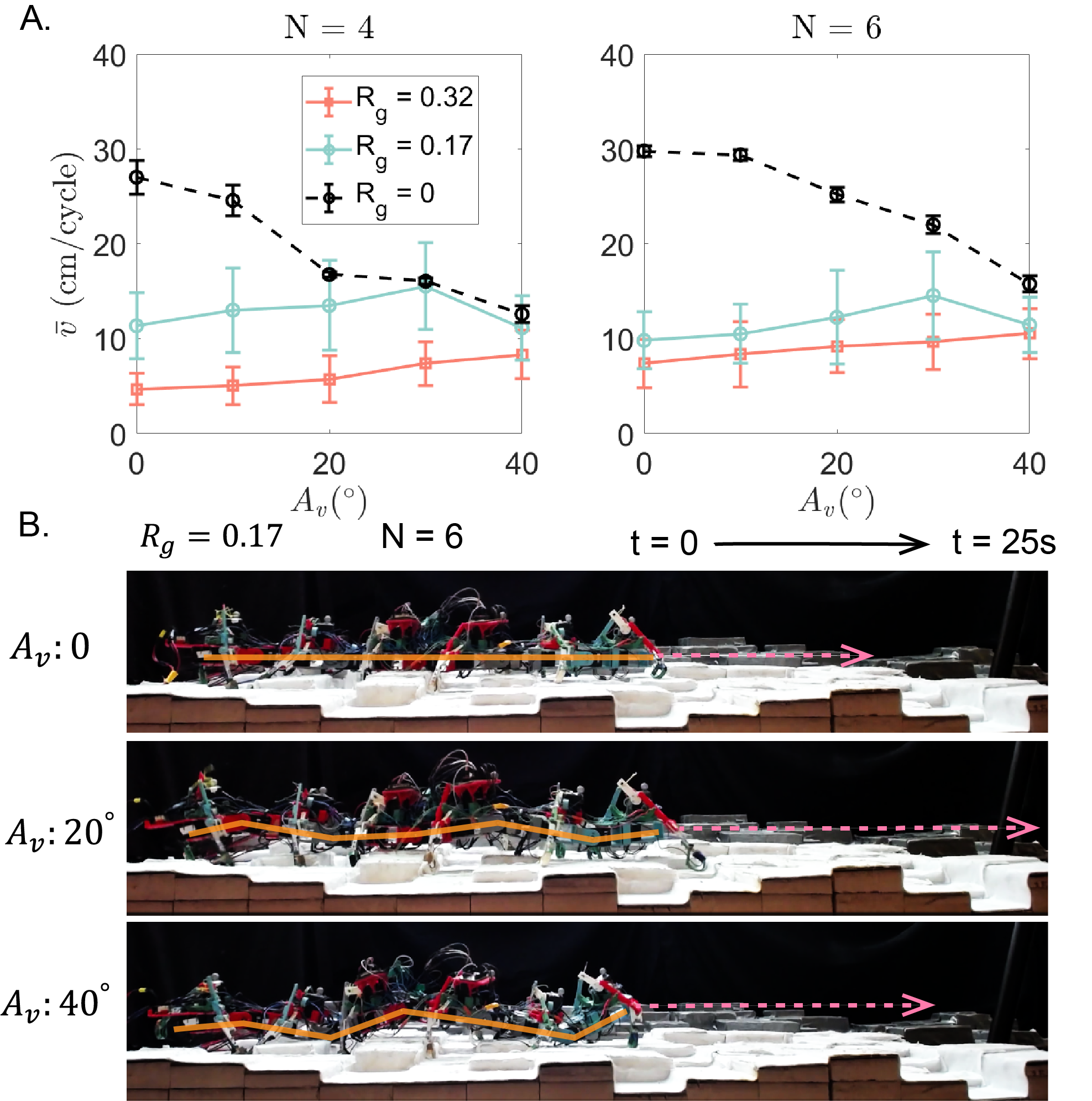}
    \caption{\textbf{Empirical evidence supporting the necessity of vertical motion modulation.} \textbf{A.} Empirical results demonstrate the relationship between the robot's speed $\bar{v}$ and the vertical amplitude $A_v$. Here, $N$ represents the number of leg pairs on the robot. For example, if $N = 4$, the robot has 8 legs. \textcolor{black}{We conducted 10 experimental trials for each data point. Error bars represent the standard deviation. }\textbf{B.} Snapshots show the differences in speed for a 12-legged robot with varying vertical amplitudes $A_v$.}
    \label{fig:10}
\end{figure}
In Section \ref{sec vertical wave}, we show how vertical body motion effectively mitigates environmental disturbances on the robot's speed. Our experimental findings, depicted in Fig. \ref{fig:10}, highlight the importance of adapting vertical motion to optimize speed in response to changes in terrain rugosity. \textcolor{black}{Specifically, the empirical data suggests that robot's speed $\bar{v}$ is reduced by increasing vertical amplitude $A_v$ on flat ground ($R_g = 0$). However, on rough terrain ($R_g=0.17 \text{ or } 0.32$), increasing $A_v$ can increase $\bar{v}$. These findings indicate that implementing feedback control of $A_v$ based on terrain rugosity could be beneficial for maximizing the robot’s forward speed. The sensitivity of the robot’s performance to temporal frequency is discussed in the Appendix. \ref{frequency sensitivity}.} 
\begin{figure*}[!t]
    \centering
    \includegraphics[width=15cm]{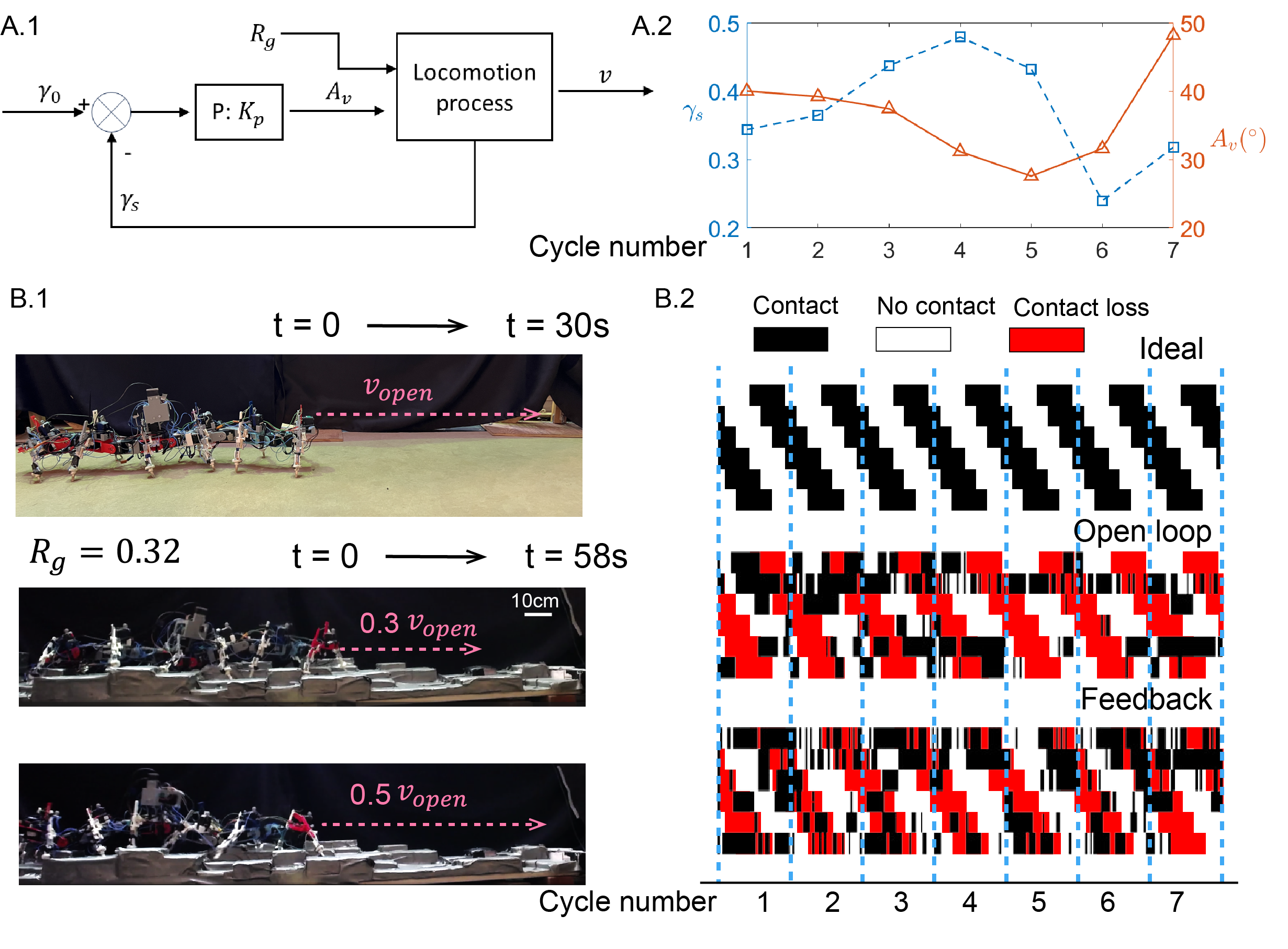}
    \caption{\textbf{Block diagram for the feedback controller and corresponding experimental results.} \textbf{A} (Left) The block diagram illustrates the feedback controller, where $\gamma_0$ represents the expected contact ratio and $\gamma_s$ is the actual contact ratio measured by the binary contact sensors. $K_p$ is the proportional gain for the linear controller. The vertical amplitude ($A_v$) for the next motion cycle is computed as $A_v = K_p (\gamma_0 - \gamma_s)$. (Right) The history of $\gamma_s$ and $A_v$ is shown for a representative trial in the feedback control experiments. \textbf{B.} (Left) Snapshots compare the performance of open-loop control and feedback control on rough terrain in terms of the average forward speed. (Right) The corresponding histories of the contact state of the left half of the robot's legs.}
    \label{fig:11}
\end{figure*}

In this section, we develop a feedback control framework that adapts vertical body motion by monitoring the foot-ground contact state and approximating terrain rugosity. To demonstrate the effectiveness of our feedback controller, we tested both our 8-legged and 12-legged robots on lab-based and outdoor terrains. We compared the performance of an open-loop controller with the feedback controller to asses how the implementation of the feedback control improves of the robot's speed.

\subsection{Vertical motion adaptation based feedback controller}

Based on the predictive analysis illustrated in Fig. \ref{fig:8}.B, optimizing $\gamma$ necessitates modulation of vertical motion. The findings in Fig. \ref{fig:9}.B reveal a correlation between the robot's velocity and the contact ratio $\gamma$, indicating higher contact ratios  lead to increased speed. Consequently, adapting vertical motion becomes crucial for optimizing the robot's traversal speed on rough terrains. 

Following this framework, we developed a linear controller (illustrated in Fig. \ref{fig:11}.A.1) to autonomously adjust the robot's vertical wave amplitude ($A_v$) when traversing rugged terrain. The onboard controller calculates the real-time contact ratio, $\gamma_s$, using data from binary contact sensors and compares $\gamma_s$ to the predetermined set contact ratio, $\gamma_0$. This comparison determine the appropriate vertical motion adjustment for the next cycle. The vertical wave amplitude is then computed as follows:

\begin{equation}
A_v (T^+) = K_p (\gamma_0-\gamma_s(T))
\end{equation}
wherein $K_p$ signifies the proportional gain for the linear controller, and $T$ and $T^+$ denote the current and next motion cycles, respectively.
\begin{figure}[!h]
    \centering
    \includegraphics[width=8.5cm]{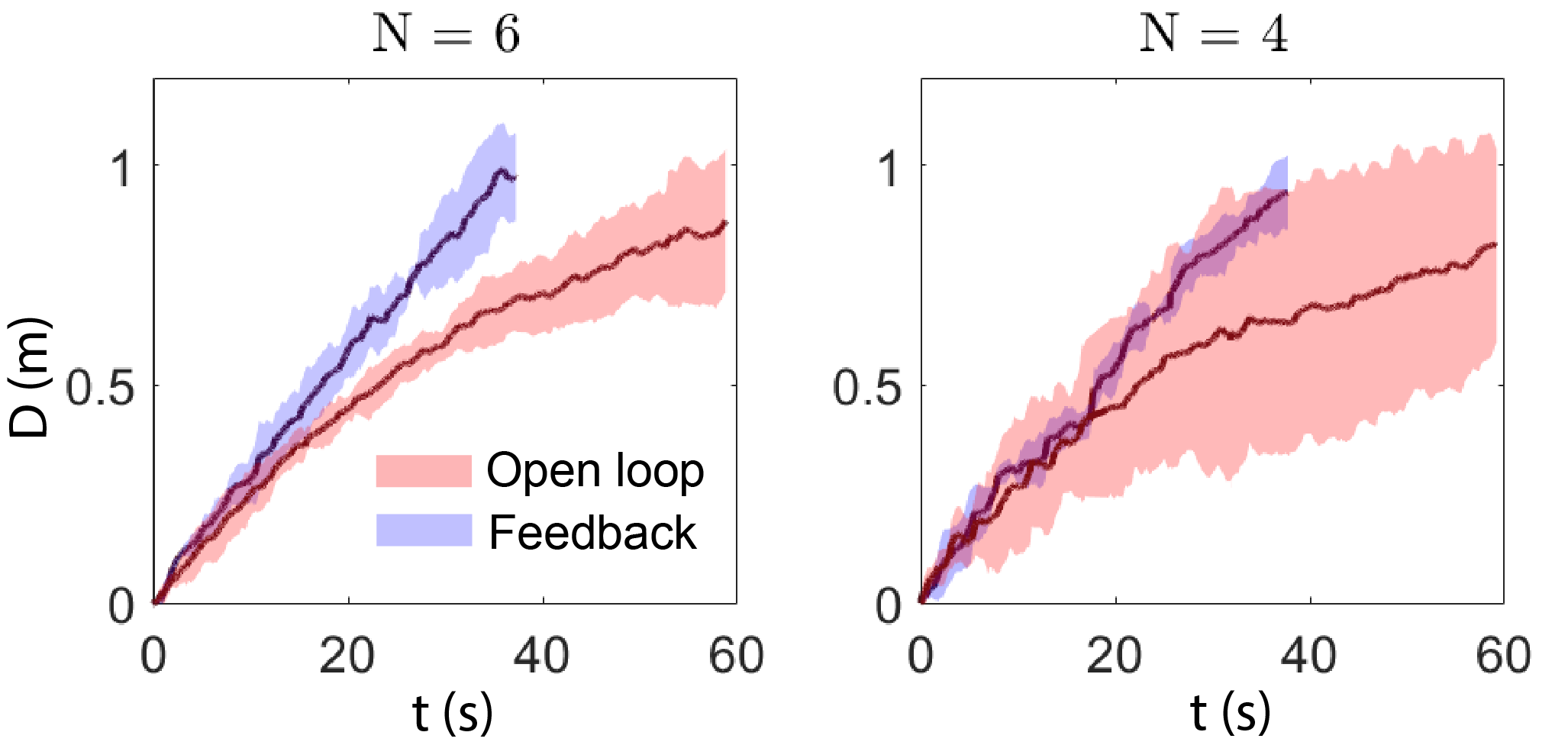}
    \caption{\textbf{Controller comparison in laboratory experiments.} Experiments were conducted on 12-legged (N = 6) and 8-legged (N = 4) robots to compare the performance of open-loop and feedback controllers. Here, D represents the forward displacement of the robots. \textcolor{black}{Each condition was tested in 10 trials, and the shaded region represents the standard deviation.}}
    \label{fig:12}
\end{figure}
\begin{figure}
    \centering
    \includegraphics[width=8.5cm]{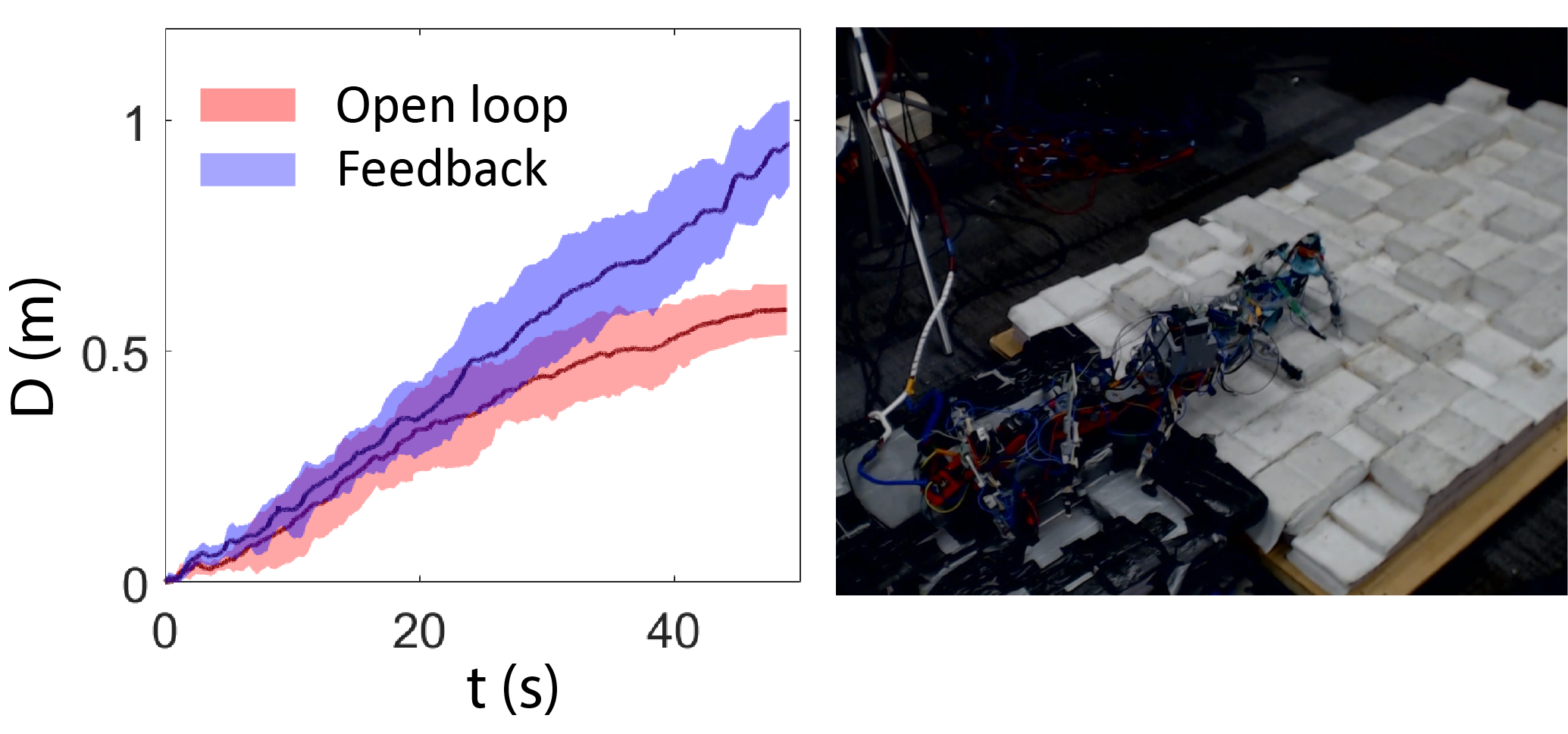}
    \caption{\textcolor{black}{\textbf{ Controller comparison in composite terrain.} The left plot shows the forward displacement over time for both open-loop and feedback controllers as the robot traverses a composite terrain consisting of two roughness levels (\(R_g = 0.17\) and \(R_g = 0.32\)). Shaded regions represent the standard deviation over five trials. The right image shows the 12-legged robot navigating the composite terrain, which features abrupt transitions in elevation and surface characteristics.}}
    \label{fig:two terrain}
\end{figure}
\begin{figure} [!h]
    \centering
    \includegraphics[width=9cm]{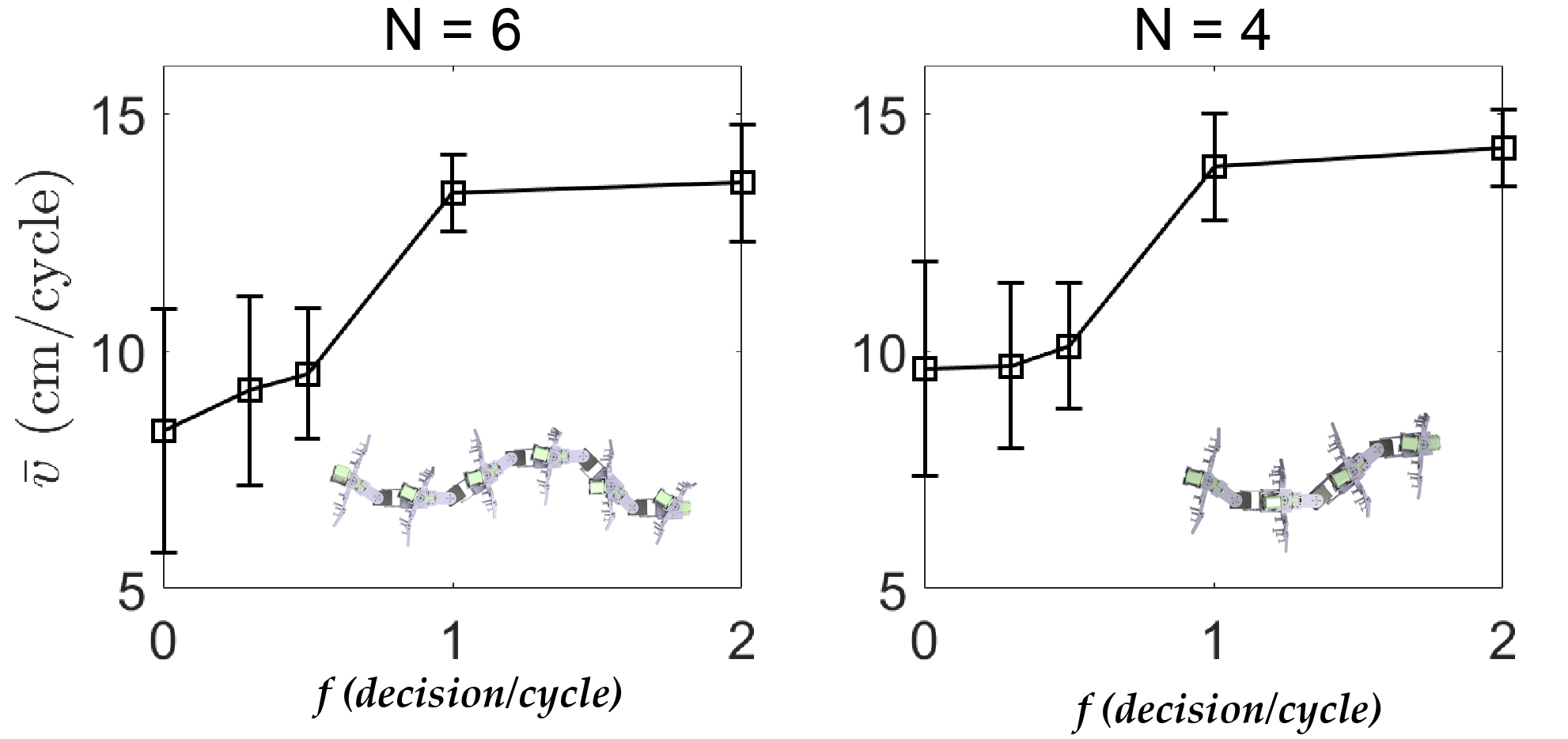}
    \caption{Experimental results offer empirical insights into how the frequency of vertical wave modulation affects the performance of the feedback controller.}
    \label{fig:13}
\end{figure}
In this feedback controller, the parameter $\gamma_s$ serves as an approximation of terrain rugosity. Since the contact ratio $\gamma$ quantifies the likelihood of the actual foot-ground contact state compared to the ideal state, rugged terrain expected to exhibit lower values of $\gamma_s$, where flatter terrains expected to yield higher values. Based on this relationship, the controller adjusts the vertical wave amplitude $A_v$ accordingly, increasing it when $\gamma_s$ is low to improve contact and reducing it when $\gamma_s$ is high.

\subsection{Lab based experiment}
The feedback controller's efficacy was assessed through experimentation conducted on laboratory-based rough terrain, specifically focusing on terrain with higher rugosity ($R_g=0.32$). Over seven motion cycles, the contact state of each leg and robot's forward displacement were recorded. \textcolor{black}{Fig. \ref{fig:11}.A} shows the temporal evolution of the measured $\gamma_s$ alongside the corresponding $A_v$ for a representative trial of the feedback control experiments. \textcolor{black}{We tuned $K_p$ from 0 to 180 degrees to obtain the optimal performance for the feedback controller. We set $K_p$ as 90 degrees for the lab-based experiments. We also conducted experiments for $K_p$ sensitivity analysis and the results can be found in Appendix. \ref{K_p sensitivity}.}

Comparing the contact states obtained using the feedback controller to those under an open-loop controller (Fig.\ref{fig:11}.B) reveals significant improvements.  The feedback controller effectively mitigates the discrepancy between the actual contact state and the ideal counterpart.
\begin{figure*}[!h]
    \centering
    \includegraphics[width=17cm]{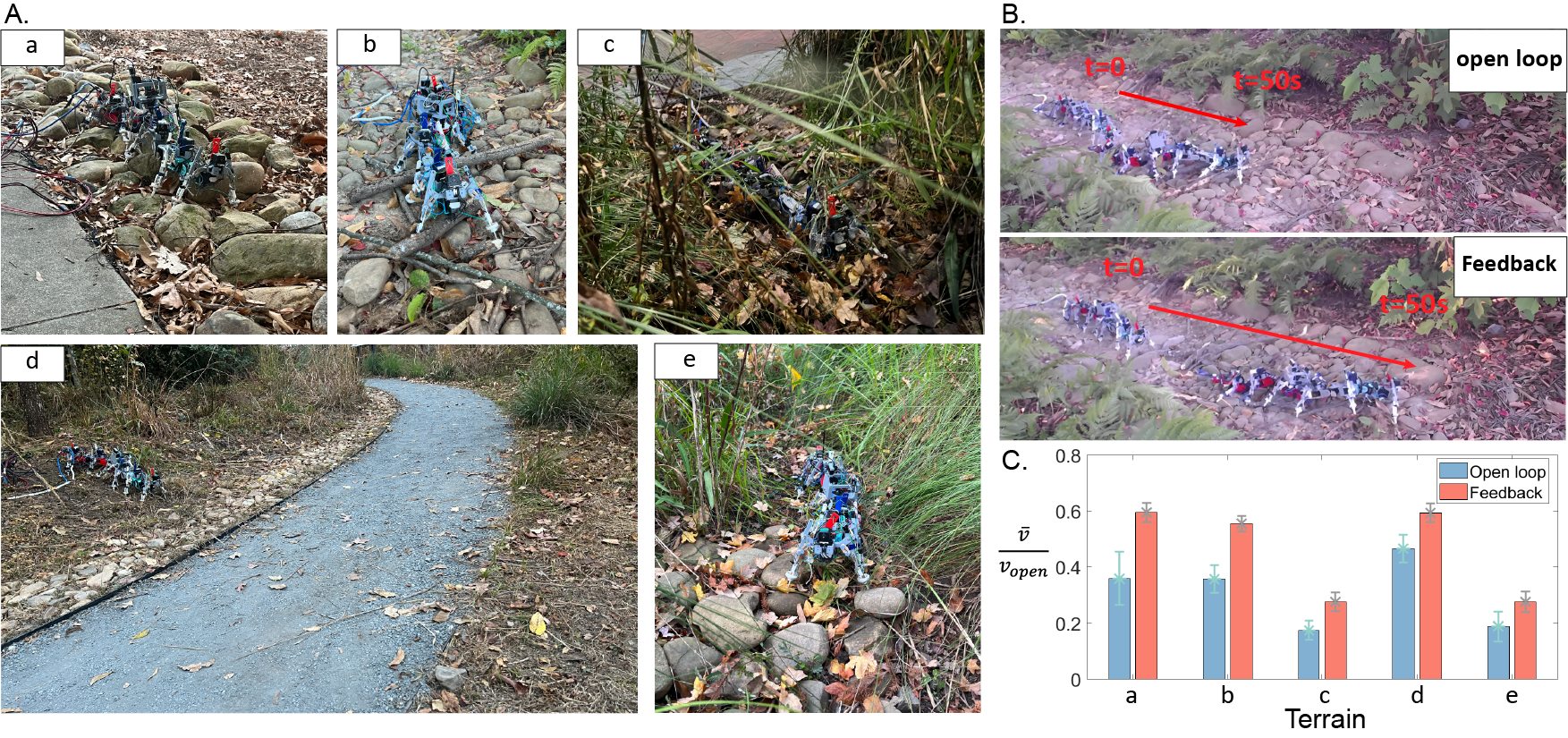}
    \caption{\textbf{Outdoor experiments testing feedback control vs. open loop in a six-segment, 12 leg robot.} \textbf{A.} Images depict the terrestrial features of five outdoor terrains. \textbf{B.} Snapshots capture representative trials of open-loop and feedback control in terrain b. \textbf{C.} Forward speed data from experiments conducted across all five outdoor terrains. $\bar{v}$ represents the average speed over a cycle of motion, while $v_{open}$ denotes the speed in open space.}
    \label{fig:14}
\end{figure*}
Figure.\ref{fig:12}.A illustrates the disparity in the robot's speed between the two controllers. Remarkably, for both the 8-legged and 12-legged robot configurations, the feedback controller not only enhances the robot's speed but also reduces speed variance. These observations highlight the effectiveness  and consistency of the feedback control mechanism in optimizing locomotion performance across various robot configurations.

\textcolor{black}{To evaluate the real-time performance of the feedback controller, we tested the robot on a composite terrain combining sections of $R_g=0.17$ and $R_g=0.32$ (Fig. \ref{fig:two terrain}). The rapid transition in terrain rugosity at the junction of these two surfaces presents additional challenges for the controller. We ran the robot for eight motion cycles on this composite terrain using both open-loop and feedback controllers, conducting five trials for each condition. As shown in Fig. \ref{fig:two terrain}, the feedback controller improved the robot’s speed by approximately 50\%, demonstrating its effectiveness even under rapidly changing terrain conditions.}

In previous feedback experiments, the robot adjusted its vertical amplitude on a per-cycle basis. To determine if its modulation frequency was optimal, we conducted additional tests on both 8-legged and 12-legged configurations with varying modulation frequencies. Our empirical findings \textcolor{black}{(Fig.\ref{fig:13})} suggest that cycle-wise vertical motion modulation emerges as the optimal approach. This method not only enhances the robot's performance on rough terrain but also ensures computational efficiency.  

\subsection{Outdoor experiment}
We tested the feedback controller's performance on five outdoor terrains with varied obstacles, including tree debris, grass, boulders, mud and rocks (Fig.\ref{fig:14}). \textcolor{black}{Terrain a consists of a mixture of robot-sized boulders, leaves, and mud. Terrain b combines robot-sized boulders with tree debris. Terrain c includes leaves, robot-sized boulders, and weeds. Terrain d features a mix of robot-sized boulders, rocks, grass, and pine straw. Terrain e contains a dense combination of highly entangled weeds and robot-sized boulders.} In these highly unpredictable environments, the feedback controller increased the robot's speed by 30-60\%, reaching up to 60\% of its maximum velocity in open areas. Additional details  and visual demonstrations of the outdoor experiments can be accessed in the Supplementary Information (SI) videos.

\section{Conclusion}
This paper presents a framework to enhance multi-legged robot locomotion. We developed a binary contact sensing system to detect foot-ground contact, forming the basis for feedback control mechanisms. Building on this, we devised probabilistic models based on stochastic leg contact states, derived from testing various vertical body wave amplitudes on terrains with different rugosities. By averaging the likelihood between actual and ideal contact states, we predict the robot's speed on these terrains, revealing how environmental disturbance impact locomotion dynamics.

Our theoretical analysis and experimental validation demonstrate the critical role of vertical body wave modulation in optimizing robot speed on rough terrain. This led to the development of a feedback control framework for automatic vertical motion modulation, which, when extensively tested in both laboratory and outdoor environments, significant improved performance performance. In laboratory trials, we observed a 50–60\% increase in robot speed and a 30–50\% reduction in speed variance compared to open-loop control. Similarly, outdoor experiments, conducted on terrains with rugosity regimes different from those in the laboratory trails, showed a 30–60\% increase in speed, with the robot reaching up to 60\% of its maximum velocity in open terrain.

\section{\textcolor{black}{Discussion and Limitations}}
\label{Discussion and limitation}
\subsection{\textcolor{black}{Model limitations}}
\textcolor{black}{In this work, we developed two probabilistic models: one capturing the correlation between the robot's speed and contact ratio, and another predicting the contact ratio based on terrain rugosity and vertical amplitude. However, there are several limitations to these models that are worth noting and addressing in future work.}

Our analysis \textcolor{black}{in Section \ref{speed correlation} for the first model} contained errors due to two key assumptions, along with other factors. First, we assumed that when a leg loses contact, the support force is lost rather than redistributed to other legs, based on the hypothesis that the force transfers to the robot’s belly. Second, we assumed the probability distribution of the slipping angle on flat terrain remains constant, though it may change on rough terrain. Additionally, sensor noise introduces errors in measuring the contact ratio, $\gamma$. To simplify calculations, we neglected the impact of collisions between the robot and the environment, despite their effect on thrust generation. These factors contributed to approximately 15\% of the data points falling outside our predictions (Fig.\ref{fig:9}.B).

\textcolor{black}{Further, We also assume that two models can be applied to robots with different numbers of legs. However, results in Fig.\ref{fig:10}A indicate that robots with fewer legs are more sensitive to terrain rugosity. This is primarily because our model assumes negligible yaw motion—an assumption that becomes less valid as the number of legs decreases. Observations show that when the head segment collides with an obstacle, yaw motion is induced, particularly in robots with fewer segments. For example, in the four-segment (eight-legged) configuration, yaw motion becomes significant, leading to lateral deviations and a more rapid reduction in forward speed as terrain rugosity increases from 0.17 to 0.32 (Fig.\ref{fig:10}).} 

\textcolor{black}{To improve predictive accuracy across different robot morphologies, future work could introduce an additional term to account for leg number, modifying the speed function from $\bar{v} = f(\gamma)$ to $\bar{v} = f(\gamma) + g(n_l)$, where $n_l$ represents the number of legs. This refinement would enhance the model’s applicability to robots with different configurations and improve its ability to predict locomotion performance across varied terrains.}

\textcolor{black}{In addition, two models in the paper rely on the contact ratio $\gamma$, without explicitly accounting for the quality of contact. In particular, the model does not differentiate between stable (firm ground) and unstable (e.g., slippery or loose) contacts, which can influence thrust generation. }

\textcolor{black}{For terrains with low friction coefficients, the gait framework used in this paper is designed to accommodate slipping during locomotion. In our previous work \cite{chong2023pnas}, we showed that thrust is governed by rate-independent Coulomb friction. During locomotion, the foot periodically slips forward and backward, balancing thrust and drag. Experimental results demonstrated that the robot’s speed (displacement per cycle) was relatively insensitive to terrain friction variations, provided there are no significant local shifts in friction. In this paper, all laboratory based experiments were conducted on surfaces with constant friction conditions. Nonetheless, we acknowledge that rapid changes in friction—such as alternating patches of mud, ice, or sand—may challenge this assumption \cite{gazzola2015gait}, and we plan to explore these conditions in future work.}

\textcolor{black}{Importantly, even with friction variability, our feedback controller is expected to enhance performance by maintaining a higher contact ratio $\gamma$, resulting in greater thrust and forward speed compared to open-loop control. This is supported by our outdoor experiments (Fig. \ref{fig:14}), where terrains (a, b, and e) included both mud and slippery boulders, and the feedback controller still achieved up to a 50\% improvement in speed.}

\textcolor{black}{For loose granular media such as sand, where the leg-ground interaction exhibits both solid- and fluid-like behavior, we recognize that a new modeling framework will be necessary, as Coulomb-based assumptions may no longer hold. Exploring this direction is part of our ongoing and future work.}

\subsection{\textcolor{black}{Control framework limitations}}
\textcolor{black}{In this paper, we developed a linear controller to improve the robot's speed on rugged terrain; however, this simple approach has certain limitations, and there remains room for further performance enhancement.}

\textcolor{black}{To support our hypothesis that vertical body undulation enhances robot performance, we assume that the heights of rough terrain follow a normal distribution — a condition that aligns with our lab-based experimental setup. However, real-world terrain is typically more diverse and complex. While we demonstrated the effectiveness of our controller in more challenging outdoor tests, future work will focus on evaluating the model across a broader range of terrain types to better capture real-world variability.}

We employed a sinusoidal traveling wave model to drive the robot's vertical body undulation. While effective in mitigating environmental disturbances through amplitude adjustments, this approach cannot generate more complex vertical body shapes. Future research will investigate the impact of more discrete vertical body shape on robot performance, focusing on how subtle variations can optimize locomotion efficiency and adaptability on complex terrains. Our recent work has produced promising results \cite{he2025climbing}.

While this study utilized a basic linear controller to enhance performance on rugged terrain, we propose that integrating the two probabilistic models with a more advanced control strategy, such as a learning-based algorithm, could significantly enhance the robot's ability to navigate complex environments. We are actively exploring this approach \cite{he2024learning}.

\subsection{\textcolor{black}{Applicability to general platforms}}
\textcolor{black}{Although the control framework is demonstrated to be effective in this paper, it remains an open question whether it is applicable to legged robots with different morphologies and dimensions.}

\textcolor{black}{Following this study, our group has developed additional multi-legged robots with varying numbers of legs \cite{he2025climbing}, body dimensions, and leg-to-body length ratios. Across these platforms, we consistently observed that vertical body undulation enhances locomotion speed, suggesting that the benefits of this mechanism are not limited to a specific morphology.
For quadrupedal and hexapod robots, we posit that introducing vertical body undulation could similarly improve performance—particularly by increasing stability on uneven or bumpy terrain. While the exact parameters may need to be tuned based on each platform’s kinematics, the underlying principle of using body undulation to mitigate environmental disturbances is expected to be broadly applicable.
}


%

\appendix
\appendices
\section*{Motion tracking for experiments}
\begin{figure}[h]
    \centering
    \includegraphics[width=8.8cm]{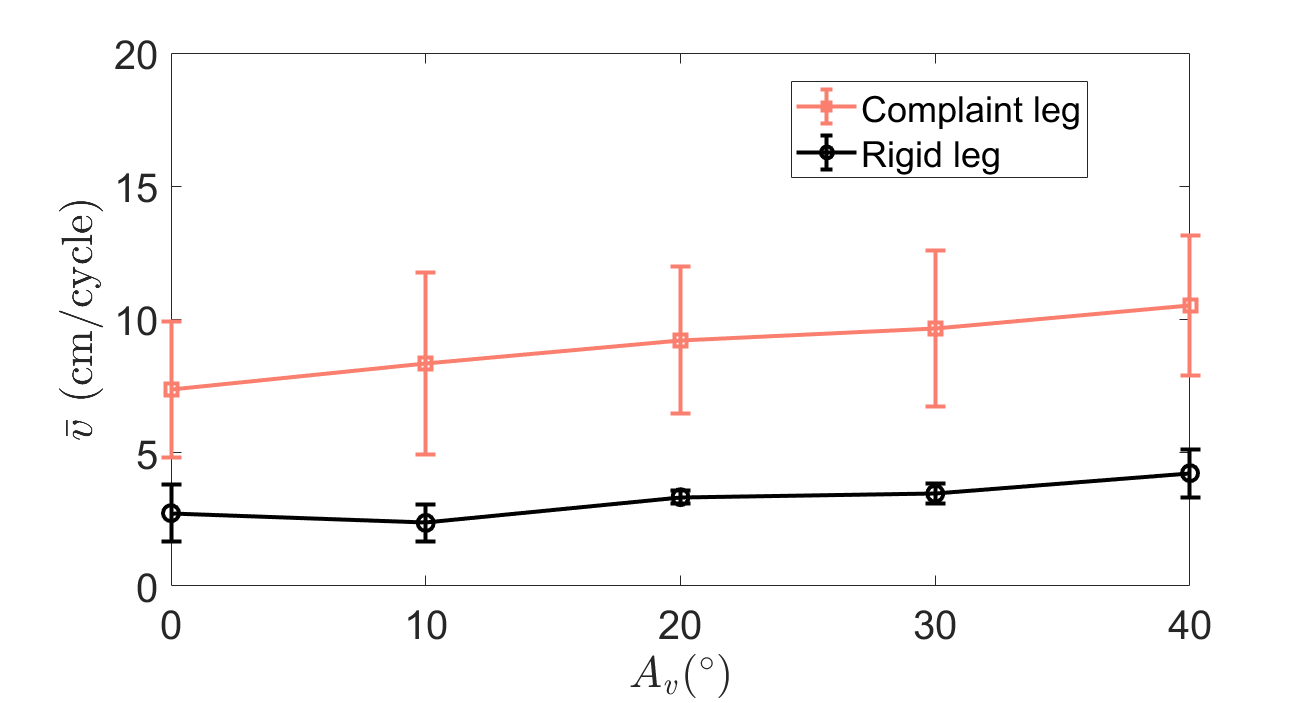}
    \caption{\textcolor{black}{\textbf{Robot performance with and without compliant legs.} The robot completed three motion cycles on rough terrain (\( R_g = 0.32 \)) using both compliant and rigid leg configurations. Across all tested vertical amplitudes, the robot equipped with compliant legs achieved more than a 200\% increase in speed compared to the rigid-legged configuration.}}
    \label{fig:rigid_vs_complaint}
\end{figure}

Reflective markers were affixed to each robot module to track its position and orientation over time. The OptiTrack motion capture system, consisting of four Prime 17W cameras recording at 360 frames per second and Motive software, tracked the markers positions within the workspace. The collected data was then analyzed using MATLAB.
\textcolor{black}{\section*{Additional experiments}
\subsection{Complaint leg test}
\label{rigid_vs_complaint}
In Section \ref{robo_model}, we introduced the compliant leg design, inspired by the design in \cite{ozkan2020systematic}, to increase robot's obstacle negotiation capability. We also conducted experiments to compare the performance of rigid leg and complaint leg on rough terrain ($R_g = 0.32$). We varied the vertical amplitude from 0 to $40^{\circ}$, with $10^{\circ}$ increment. For each condition, we performed 5 trials, and in each trial, the robot completed 3 motion cycles. The plot below presents the results of these experiments.  As Fig. \ref{fig:rigid_vs_complaint} shows, the complaint leg increases the forward speed of the robot by around 200\%. }
\textcolor{black}{
\subsection{Gait stability at higher speeds}\label{frequency sensitivity}}

\textcolor{black}{
In the lab-based and outdoor experiments presented in this paper, the robot was operated at a temporal frequency of 1/6 Hz (6 seconds per motion cycle). To further investigate the effect of higher speeds, we conducted additional experiments at increased frequencies: 1/3 Hz (3 seconds per cycle) and 1 Hz (1 second per cycle), on rough terrain with \( R_g = 0.17 \). We tested three different vertical amplitudes: \(10^{\circ}\), \(20^{\circ}\), and \(30^{\circ}\). For each condition, we conducted five trials, with the robot completing three motion cycles per trial.}

\begin{figure}[h]
    \centering
    \includegraphics[width=9cm]{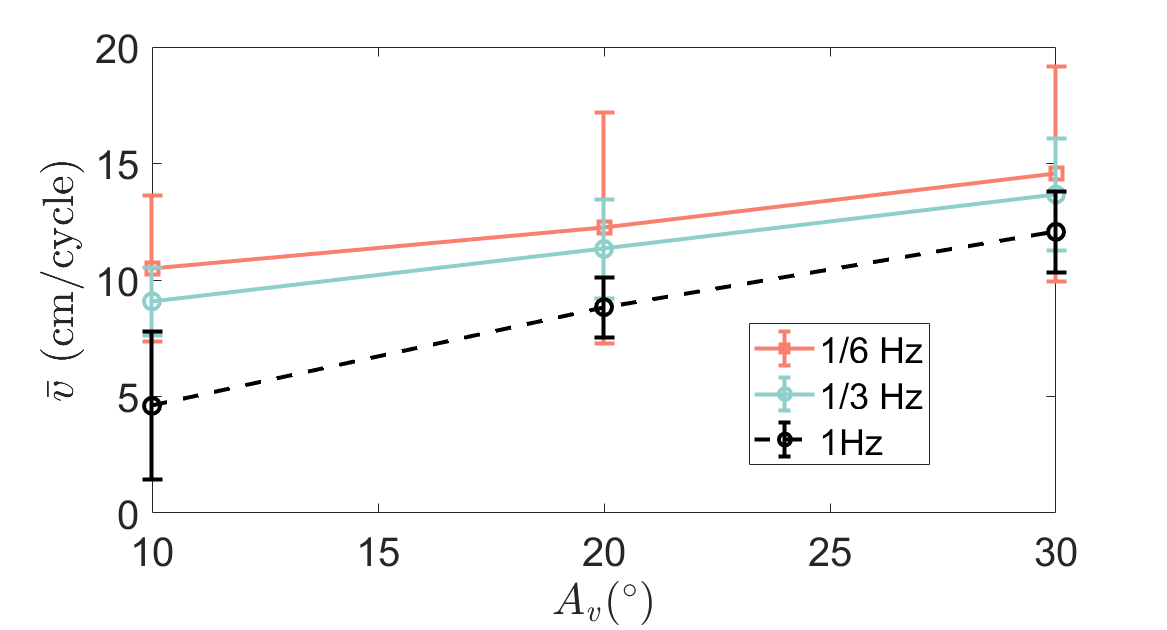}
    \caption{\textcolor{black}{\textbf{Gait performance at higher temporal frequencies.} The plot shows the average forward displacement per cycle \(\bar{v}\) as a function of vertical body wave amplitude \(A_v\), evaluated at three temporal frequencies: 1/6 Hz (6 seconds per cycle), 1/3 Hz (3 seconds per cycle), and 1 Hz (1 second per cycle), on rough terrain with \(R_g = 0.17\). Each data point represents the average over five trials, with error bars indicating standard deviation. Results show that gait performance remains stable when increasing frequency from 1/6 Hz to 1/3 Hz. At 1 Hz, performance drops significantly at low amplitudes but remains relatively high when larger vertical amplitudes are used, suggesting that greater body undulation helps maintain stability and efficiency at higher speeds.}}
    \label{fig:temporal freq}
\end{figure}

\textcolor{black}{The results (Fig.~\ref{fig:temporal freq}) show that gait performance remains consistent when the temporal frequency increases from 1/6 Hz to 1/3 Hz. However, at 1 Hz, the robot's displacement per cycle decreases to approximately 50\% of the 1/6 Hz value for the smallest vertical amplitude (\(10^{\circ}\)), and to about 85\% for the highest vertical amplitude (\(30^{\circ}\)). These findings suggest that larger vertical amplitudes help maintain gait performance at higher speeds. We plan to further investigate this phenomenon in future work.
}
\textcolor{black}{\subsection{$K_p$ sensitivity analysis}
\label{K_p sensitivity}
To evaluate the impact of the proportional gain parameter $k_p$ in our feedback controller, we conducted a series of sensitivity experiments. Recall that the vertical body wave amplitude is computed as,  
$$
A_v = k_p (\gamma_0 - \gamma_s),
$$  
where $\gamma_0$ is the nominal contact ratio and $\gamma_s$ is the sensed contact ratio.}

\textcolor{black}{We varied $k_p$ from $0^\circ$ to $180^\circ$ in $30^\circ$ increments. For each value, the robot completed five motion cycles per trial, and each condition was tested across five repeated trials to ensure robustness.}

\textcolor{black}{When $k_p$ is small, the resulting vertical amplitude $A_v$ remains near zero, and the robot's behavior closely resembles a gait without vertical undulation. Conversely, at higher $k_p$ values (approaching $180^\circ$), even a small difference between $\gamma_0$ and $\gamma_s$ produces a large $A_v$. Since we cap the vertical amplitude at $50^\circ$ to prevent self-collision, setting $k_p$ near $180^\circ$ effectively causes the robot to operate at its maximum vertical amplitude throughout the motion cycle.}

\textcolor{black}{The figure below illustrates how the average forward speed per cycle varies with $k_p$. As shown, speed increases with $k_p$, reaching a peak around $k_p = 90^\circ$, after which it slightly declines. These results underscore the importance of gain selection and demonstrate the sensitivity of the controller to the $k_p$ parameter. 
}
\begin{figure}[h]
    \centering
    \includegraphics[width=9cm]{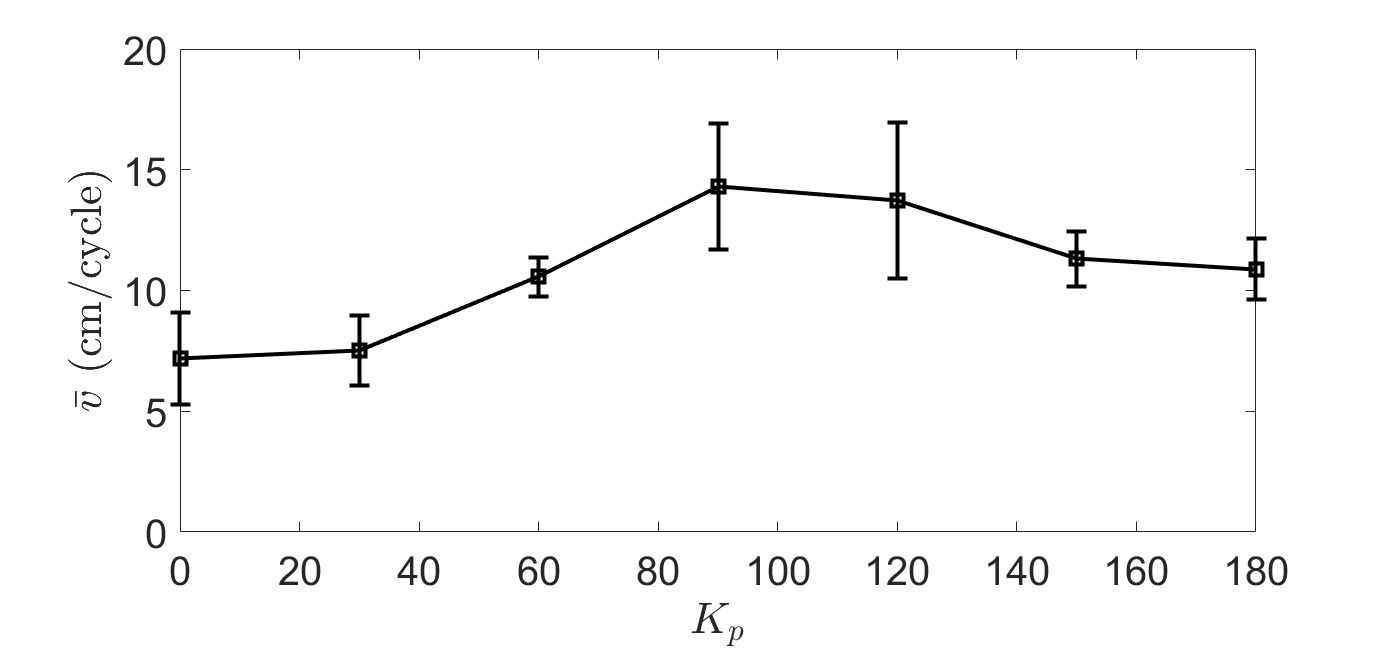}
    \caption{\textcolor{black}{\textbf{Sensitivity analysis of \(K_p\).} The plot shows the average forward speed per cycle \(\bar{v}\) as a function of \(K_p\), which determines the vertical body undulation amplitude according to \(A_v = K_p (\gamma_0 - \gamma_s)\). Each data point represents the mean speed over five trials, with each trial consisting of three motion cycles. Error bars indicate the standard deviation across trials. The results show that performance improves with increasing \(K_p\), peaking around \(K_p = 90^\circ\), after which the speed slightly declines. This trend highlights the importance of appropriate gain tuning in the feedback controller.}}
    \label{fig:kp_sensitivity}
\end{figure}

\section*{Acknowledgment}
We would like to thank Daniel Irvine and Jeff Aguilar for helpful discussions. We are grateful for funding from NSF STTR AWD-2335553, Army Research Office Grant AWD-001950, Georgia Research Alliance (GRA) AWD-005494, GA AIM AWD-004173 and the Dunn Family Professorship.


\ifCLASSOPTIONcaptionsoff
  \newpage
\fi



\bibliographystyle{IEEEtran}
\bibliography{bibtex/bib/IEEEabrv,bibtex/bib/IEEEexample}
\end{document}